\documentclass[journal]{IEEEtran}
\IEEEoverridecommandlockouts

\usepackage{times}
\usepackage{epsfig}
\usepackage{graphicx}
\usepackage{amsmath}
\usepackage{amssymb}
\usepackage{subfigure}
\usepackage{colortbl}
\usepackage{multirow}
\usepackage{soul}
\usepackage{adjustbox}
\usepackage{array}
\usepackage{setspace}
\usepackage{lscape}
\usepackage{mathtools}
\usepackage{hyperref}
\usepackage{adjustbox}
\usepackage{cite}
\usepackage[super]{nth}
\usepackage[dvipsnames]{xcolor}

\newcommand{\printfnsymbol}[1]{%
  \textsuperscript{*}%
}

\begin{document}
%

\title{DeepFakes and Beyond: A Survey of \\ Face Manipulation and Fake Detection}

%
%
%

%

\author{Ruben Tolosana, Ruben Vera-Rodriguez, Julian Fierrez, Aythami Morales and Javier Ortega-Garcia\\Biometrics and Data Pattern Analytics - BiDA Lab, Universidad Autonoma de Madrid, Spain \\ \{ruben.tolosana, ruben.vera, julian.fierrez, aythami.morales, javier.ortega\}@uam.es}



%



\maketitle

\begin{abstract}
The free access to large-scale public databases, together with the fast progress of deep learning techniques, in particular Generative Adversarial Networks, have led to the generation of very realistic fake content with its corresponding implications towards society in this era of fake news. 

This survey provides a thorough review of techniques for manipulating face images including DeepFake methods, and methods to detect such manipulations. In particular, four types of facial manipulation are reviewed: \textit{i)} entire face synthesis, \textit{ii)} identity swap (DeepFakes), \textit{iii)} attribute manipulation, and \textit{iv)} expression swap. For each manipulation group, we provide details regarding manipulation techniques, existing public databases, and key benchmarks for technology evaluation of fake detection methods, including a summary of results from those evaluations. Among all the aspects discussed in the survey, we pay special attention to the latest generation of DeepFakes, highlighting its improvements and challenges for fake detection. 

In addition to the survey information, we also discuss open issues and future trends that should be considered to advance in the field.
\end{abstract}


%
\IEEEpeerreviewmaketitle

\section{Introduction}
\IEEEPARstart{F}{ake} images and videos including facial information generated by digital manipulation, in particular with DeepFake methods~\cite{korshunov2018deepfakes}, have become a great public concern recently~\cite{TED_news_concerns,BBC_doubleVideos}. The very popular term ``DeepFake" is referred to a deep learning based technique able to create fake videos by swapping the face of a person by the face of another person. This term was originated after a Reddit user named ``deepfakes" claimed in late 2017 to have developed a machine learning algorithm that helped him to transpose celebrity faces into porn videos~\cite{BBC_originDeepFake}. In addition to fake pornography, some of the more harmful usages of such fake content include fake news, hoaxes, and financial fraud. As a result, the area of research traditionally dedicated to general media forensics~\cite{swaminathan2008digital,farid2009image,stamm2010forensic,rocha2011vision,milani2012overview,piva2013overview,korus2017digital}, is being invigorated and is now dedicating growing efforts for detecting facial manipulation in image and video~\cite{rossler2019faceforensics++}. Part of these renewed efforts in fake face detection are built around past research in biometric anti-spoofing~\cite{galbally14reviewAntispoofingFace,2015_ISPM_PAs,handbook_Julian} and modern data-driven deep learning~\cite{2019_Arxiv_GANRemoval_Tolosana,Jain2019facialManipulation}. The growing interest in fake face detection is demonstrated through the increasing number of workshops in top conferences~\cite{CVPR_workshop,ICML_competition,ACMMMM_tutorial,WACV_workshop,ICPR_workshop}, international projects such as MediFor funded by the Defense Advanced Research Project Agency (DARPA), and competitions such as the recent Media Forensics Challenge (MFC2018)\footnote{\url{https://www.nist.gov/itl/iad/mig/media-forensics-challenge-2018}} and the Deepfake Detection Challenge (DFDC)\footnote{\url{https://deepfakedetectionchallenge.ai/}} launched by the National Institute of Standards and Technology (NIST) and Facebook, respectively.

Traditionally, the number and realism of facial manipulations have been limited by the lack of sophisticated editing tools, the domain expertise required, and the complex and time-consuming process involved. For example, an early work in this topic~\cite{VideoRewrite97} was able to modify the lip motion of a person speaking using a different audio track, by making connections between the sounds of the audio track and the shape of the subject's face. However, from these early works up to date, many things have rapidly evolved in the last years. Nowadays, it is becoming increasingly easy to automatically synthesise non-existent faces or manipulate a real face of one person in an image/video, thanks to: \textit{i)} the accessibility to large-scale public data, and \textit{ii)} the evolution of deep learning techniques that eliminate many manual editing steps such as Autoencoders (AE) and Generative Adversarial Networks (GAN)~\cite{autoencoder_ICLR,goodfellow2014generative}. As a result, open software and mobile application such as ZAO\footnote{\url{https://apps.apple.com/cn/app/id1465199127}} and FaceApp\footnote{\url{https://apps.apple.com/gb/app/faceapp-ai-face-editor/id1180884341}} have been released opening the door to anyone to create fake images and videos, without any experience in the field needed. 

\begin{figure*}[t]
\begin{center}
   \includegraphics[width=0.96\linewidth]{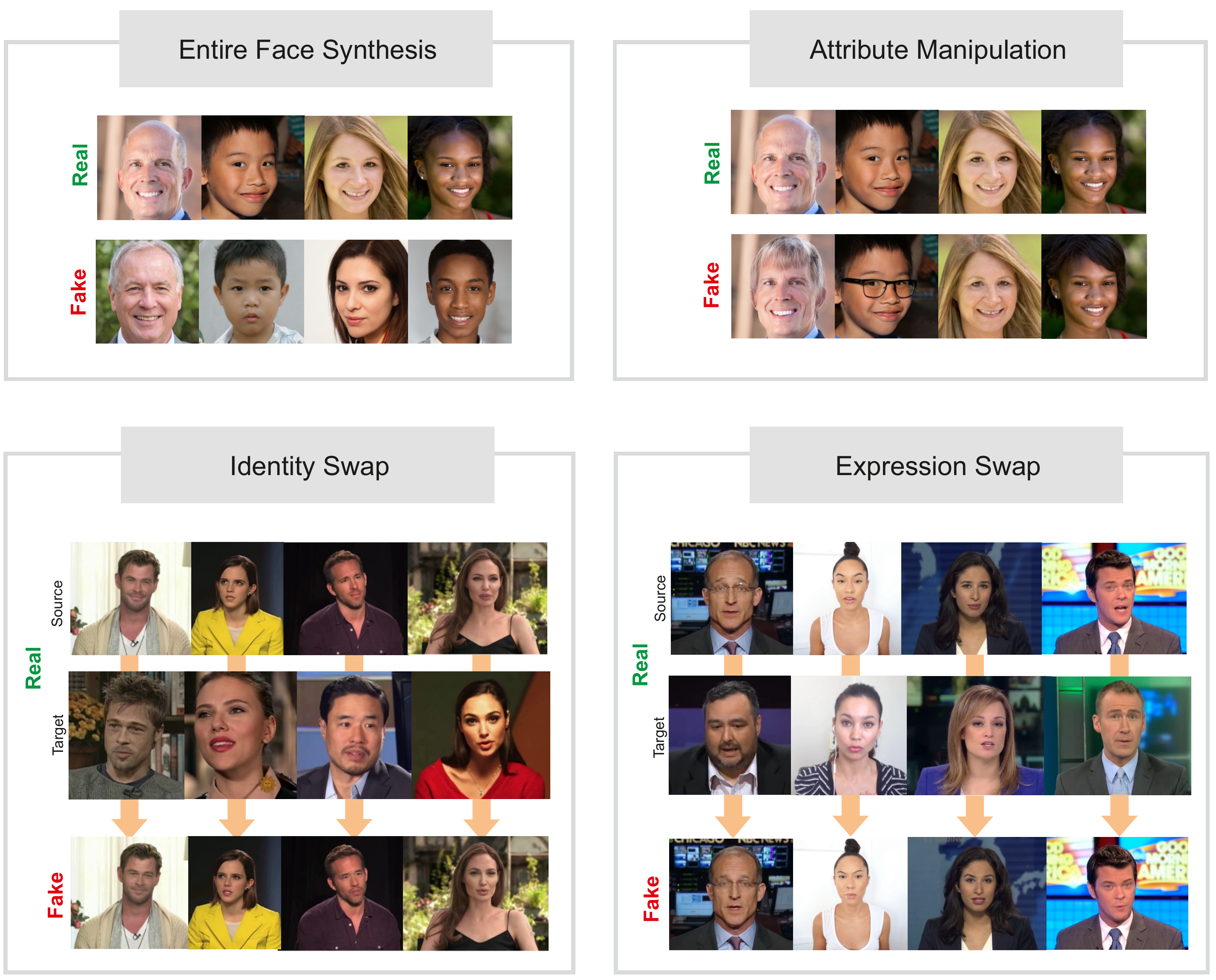}
\end{center}
   \caption{Real and fake examples of each facial manipulation group. For \textit{Entire Face Synthesis}, real images are extracted from \url{http://www.whichfaceisreal.com/} and fake images from \url{https://thispersondoesnotexist.com}. For \textit{Identity Swap}, face images are extracted from Celeb-DF database~\cite{li2019celebdf}. For \textit{Attribute Manipulation}, real images are extracted from \url{http://www.whichfaceisreal.com/} and fake images are generated using FaceApp. Finally, for \textit{Expression Swap}, images are extracted from FaceForensics++~\cite{rossler2019faceforensics++}.}
\label{fig:abstract_figure}
\end{figure*}

In response to those increasingly sophisticated and realistic manipulated content, large efforts are being carried out by the research community to design improved methods for face manipulation detection. Traditional fake detection methods in media forensics have been commonly based on: \textit{i)} in-camera fingerprints, the analysis of the intrinsic fingerprints introduced by the camera device, both hardware and software, such as the optical lens~\cite{yerushalmy2011digital}, colour filter array and interpolation~\cite{popescu2005exposing,cao2009accurate}, and compression~\cite{lin2009fast,chen2011detecting}, among others, and \textit{ii)} out-camera fingerprints, the analysis of the external fingerprints introduced by editing software, such as copy-paste or copy-move different elements of the image~\cite{amerini2011sift,cozzolino2015splicebuster}, reduce the frame rate in a video~\cite{gironi2014video,wu2014exposing,speed_video_2020}, etc. However, most of the features considered in traditional fake detection methods are highly dependent on the specific training scenario, being therefore not robust against unseen conditions~\cite{farid2009image,rocha2011vision,2019_Arxiv_GANRemoval_Tolosana}. This is of special importance in the era we live in as most media fake content is usually shared on social networks, whose platforms automatically modify the original image/video, for example, through compression and resize operations~\cite{rossler2019faceforensics++}.

This survey provides an in-depth review of digital manipulation techniques applied to facial content due to the large number of possible harmful applications, e.g., the generation of fake news that would provide misinformation in political elections and security threats~\cite{allcott2017social,lazer2018science}. Specifically, we cover four types of manipulations: \textit{i)} entire face synthesis, \textit{ii)} identity swap, \textit{iii)} attribute manipulation, and \textit{iv)} expression swap. These four main types of face manipulation are well established by the research community, receiving most attention in the last few years. Besides, we also review in this survey some other challenging and dangerous face manipulation techniques that are not so popular yet like face morphing.

Finally, for completeness, we would like to highlight other recent surveys in the field. In~\cite{kietzmann2020deepfakes}, the authors cover the topic of DeepFakes from a general perspective, proposing the R.E.A.L framework to manage DeepFake risks. In addition, Verdoliva has recently surveyed in~\cite{verdoliva2020media} traditional manipulation and fake detection approaches considered in general media forensics, and also the latest deep learning techniques. The present survey complements~\cite{kietzmann2020deepfakes} and~\cite{verdoliva2020media} with a more detailed review of each facial manipulation group, including manipulation techniques, existing public databases, and key benchmarks for technology evaluation of fake detection methods, including a summary of results from those evaluations. In addition, we pay special attention to the latest generation of DeepFakes, highlighting its improvements and challenges for fake detection.

The remainder of the article is organised as follows. We first provide in Sec.~\ref{sec_typesManipulations} a general description of different types of facial manipulation. Then, from Sec.~\ref{entire_face_synthesis} to Sec.~\ref{facial_expression} we describe the key aspects of each type of facial manipulation including public databases for research, detection methods, and benchmark results. Sec.~\ref{others_research_lines}
focuses on other interesting types of face manipulation techniques not covered in previous sections. Finally, we provide in Sec.~\ref{conclusions} our concluding remarks, highlighting open issues and future trends.

\section{Types of Facial Manipulations}\label{sec_typesManipulations}
Facial manipulations can be categorised in four main different groups regarding the level of manipulation. Fig.~\ref{fig:abstract_figure} graphically summarises each facial manipulation group. A description of each of them is provided below, from higher to lower level of manipulation:

\begin{itemize}
\item \textbf{Entire Face Synthesis:} this manipulation creates entire non-existent face images, usually through powerful GAN, e.g., through the recent StyleGAN approach proposed in~\cite{Karras_2019_CVPR}. These techniques achieve astonishing results, generating high-quality facial images with a high level of realism. Fig.~\ref{fig:abstract_figure} shows some examples for entire face synthesis generated using StyleGAN\footnote{\url{https://thispersondoesnotexist.com}}. This manipulation could benefit many different sectors such as the video game and 3D-modelling industries, but it could also be used for harmful applications such as the creation of very realistic fake profiles in social networks in order to generate misinformation.
\item \textbf{Identity Swap:} this manipulation consists of replacing the face of one person in a video with the face of another person. Two different approaches are usually considered: \textit{i)} classical computer graphics-based techniques such as FaceSwap\footnote{\url{https://github.com/MarekKowalski/FaceSwap}}, and \textit{ii)} novel deep learning techniques known as DeepFakes\footnote{\url{https://github.com/deepfakes/faceswap}}, e.g., the recent ZAO mobile application. Very realistic videos of this type of manipulation can be seen on Youtube\footnote{\url{https://www.youtube.com/watch?v=UlvoEW7l5rs}}. This type of manipulation could benefit many different sectors, in particular the film industry. However, in the other side, it could also be used for bad purposes such as the creation of celebrity pornographic videos, hoaxes, and financial fraud, among many others.  
\item \textbf{Attribute Manipulation:} this manipulation, also known as face editing or face retouching, consists of modifying some attributes of the face such as the colour of the hair or the skin, the gender, the age, adding glasses, etc~\cite{2018_TIFS_SoftWildAnno_Sosa}. This manipulation process is usually carried out through GAN such as the StarGAN approach proposed in~\cite{choi2018stargan}. One example of this type of manipulation is the popular FaceApp mobile application. Consumers could use this technology to try on a broad range of products such as cosmetics and makeup, glasses, or hairstyles in a virtual environment.
\item \textbf{Expression Swap:} this manipulation, also known as face reenactment, consists of modifying the facial expression of the person. Although different manipulation techniques are proposed in the literature, e.g., at image level through popular GAN architectures~\cite{Liu_2019_CVPR}, in this group we focus on the most popular techniques Face2Face and NeuralTextures \cite{thies2016face2face,thies2019deferred}, which replaces the facial expression of one person in a video with the facial expression of another person. This type of manipulation could be used with serious consequences, e.g., the popular video of Mark Zuckerberg saying things he never said\footnote{\url{https://www.bbc.com/news/technology-48607673}}.
\end{itemize}

\begin{table}[t]
\centering
\caption{\textbf{Entire Face Synthesis:} Publicly available databases.}
\vspace{3mm}
\label{table:databases_entireFace}
\scalebox{0.8}{
\begin{tabular}{ccc}
\textbf{Database}                                                                  & \textbf{Real Images} & \textbf{Fake Images}                                                         \\ \hline
\begin{tabular}[c]{@{}c@{}}100K-Generated-Images (2019)\\ \cite{Karras_2019_CVPR}\end{tabular} & -                    & 100,000 (StyleGAN) \\ \hline
\begin{tabular}[c]{@{}c@{}}100K-Faces (2019)\\ \cite{100kfaces}\end{tabular} & -                    & 100,000 (StyleGAN)                                                           \\ \hline
\begin{tabular}[c]{@{}c@{}}DFFD (2020)\\ \cite{Jain2019facialManipulation}\end{tabular}                  & -                    & \begin{tabular}[c]{@{}c@{}}100,000 (StyleGAN)\\ 200,000 (ProGAN)\end{tabular} \\ \hline
\begin{tabular}[c]{@{}c@{}}iFakeFaceDB (2020)\\ \cite{2019_Arxiv_GANRemoval_Tolosana}\end{tabular} & -                    &  \begin{tabular}[c]{@{}c@{}}250,000 (StyleGAN)\\ 80,000 (ProGAN)\end{tabular}
\end{tabular}
}
\end{table}

\section{Entire Face Synthesis}\label{entire_face_synthesis}

\subsection{Manipulation Techniques and Public Databases}\label{subsec:entireFace_databases}
This manipulation creates entire non-existent face images. Table~\ref{table:databases_entireFace} summarises the main publicly available databases for research on detection of image manipulation techniques relying on entire face synthesis. Four different databases are of relevance here, all of them based on the same GAN architectures: ProGAN~\cite{pgan} and StyleGAN~\cite{Karras_2019_CVPR}. It is interesting to remark that each fake image may be characterised by a specific GAN fingerprint just like natural images are identified by a device-based fingerprint (i.e., PRNU). In fact, these fingerprints seem to be dependent not only of the GAN architecture, but also of the different instances of it~\cite{marra2019gans,albright_CVPR2019,richa_vatsa_2020}.

In addition, as indicated in Table~\ref{table:databases_entireFace}, it is important to note that the four mentioned databases only contain fake images generated using the GAN architectures discussed. In order to perform fake detection experiments on this manipulation group, researchers need to obtain real face images from other public databases such as CelebA~\cite{celeba}, FFHQ~\cite{Karras_2019_CVPR}, CASIA-WebFace~\cite{yi2014learning}, and VGGFace2~\cite{cao2018vggface2}, among others.

We provide next a description of each public database. In~\cite{Karras_2019_CVPR}, Karras \textit{et al.} released a set of 100,000 synthetic face images, named 100K-Generated-Images\footnote{\url{https://github.com/NVlabs/stylegan}}. This database was generated using their proposed StyleGAN architecture, which was trained using the FFHQ dataset~\cite{Karras_2019_CVPR}. StyleGAN is an improved version of their previous popular approach ProGAN, which introduced a new training methodology based on improving both generator and discriminator progressively. StyleGAN proposes an alternative generator architecture that leads to an automatically learned, unsupervised separation of high-level attributes (e.g., pose and identity when trained on human faces) and stochastic variation in the generated images (e.g., freckles, hair), and it enables intuitive, scale-specific control of the synthesis. 

Another public database is 100K-Faces~\cite{100kfaces}. This database contains 100,000 synthetic images generated using StyleGAN. In this database, contrary to the 100K-Generated-Images database, the StyleGAN network was trained using around 29,000 photos from 69 different models, considering face images from a more controlled scenario (e.g., with a flat background). Thus, no strange artifacts created by the StyleGAN are included in the background of the images. 

Recently, Dang \textit{et al.} introduced in~\cite{Jain2019facialManipulation} a new database named Diverse Fake Face Dataset (DFFD). Regarding the entire face synthesis manipulation, the authors created 100,000 and 200,000 fake images through the pre-trained ProGAN and StyleGAN models, respectively. 

\begin{figure}[tb]
\centering
\subfigure[Fake]{\label{fig:loss_TPDNE}
\includegraphics[width=0.35\linewidth]{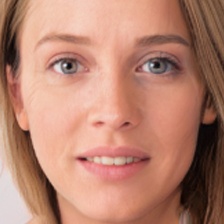}}
\subfigure[Fake after GANprintR]{\label{fig:loss_100KFaces}
\includegraphics[width=0.35\linewidth]{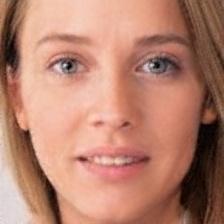}}
\caption{Examples of a fake image created using StyleGAN and its improved version after removing the GAN-fingerprint information with GANprintR~\cite{2019_Arxiv_GANRemoval_Tolosana}.} 
\label{fig:gan_fingerprint}
\end{figure}

Finally, Neves \textit{et al.} presented in~\cite{2019_Arxiv_GANRemoval_Tolosana} the iFakeFaceDB database. This database comprises 250,000 and 80,000 synthetic face images created with StyleGAN and ProGAN, respectively. As an additional feature in comparison to previous databases, and in order to hinder fake detectors, in this database the fingerprints produced by the GAN architectures were removed through an approach named GANprintR (GAN fingerprint Removal), while keeping very realistic appearance. Fig.~\ref{fig:gan_fingerprint} shows an example of a fake image directly generated with StyleGAN and its improved version after removing the GAN-fingerprint information. As a result of the GANprintR step, iFakeFaceDB presents a higher challenge for advanced fake detectors compared with the other databases.

\subsection{Manipulation Detection}\label{EntireFaceDet}
Different studies have recently evaluated the difficulty of detecting whether faces are real of artificially generated. Table~\ref{table:relatedWorks_entireFaceSynthesis} shows a comparison of the most relevant approaches in this area. For each study, we include information related to the method, classifiers, best performance, and databases considered. We highlight in \textbf{bold} the best results achieved for each public database. It is important to remark that in some cases, different evaluation metrics are considered, e.g., Area Under the Curve (AUC) or Equal Error Rate (EER), which complicates the comparison among the studies.

\begin{table*}[t]
\centering
\caption{\textbf{Entire Face Synthesis:} Comparison of different state-of-the-art detection approaches. The best results achieved for each public database are remarked in \textbf{bold}. Results in \textit{italics} indicate that they were not provided in the original work. \hspace{\textwidth} AUC = Area Under the Curve, Acc. = Accuracy, EER = Equal Error Rate. }
\vspace{3mm}
\label{table:relatedWorks_entireFaceSynthesis}
\scalebox{0.78}{
\begin{tabular}{ccccc}
\textbf{Study}                                                                     & \textbf{Method}                                                       & \textbf{Classifiers}      & \textbf{Best Performance} & \textbf{Databases (Generation)}                                                                \\ \hline
\begin{tabular}[c]{@{}c@{}}McCloskey and Albright (2018)\\ \cite{mccloskey2018detecting}\end{tabular} & GAN-Pipeline Features                                                  & SVM                       & AUC = 70.0\%                & NIST MFC2018                                                                      \\ \hline
\begin{tabular}[c]{@{}c@{}}Wang \textit{et al.} (2019)\\ \cite{wang2019fakespotter}\end{tabular}          & \begin{tabular}[c]{@{}c@{}}GAN-Pipeline Features \end{tabular} & SVM                       & Acc. = 84.7\%            & \begin{tabular}[c]{@{}c@{}}Own \\(InterFaceGAN, StyleGAN)\end{tabular}  \\ \hline 

\begin{tabular}[c]{@{}c@{}}Guarnera \textit{et al.} (2020)\\ \cite{convolutional_traces_2020}\end{tabular}     & GAN-Pipeline Features        & \textit{k}-NN, SVM, LDA & Acc. = 99.81\%                                                                        & \begin{tabular}[c]{@{}c@{}}Own\\ (AttGAN, GDWCT, \\StarGAN, StyleGAN, StyleGAN2)\end{tabular} \\  \hline \hline

\begin{tabular}[c]{@{}c@{}}Nataraj \textit{et al.} (2019)\\ \cite{nataraj2019detecting}\end{tabular}           & Steganalysis Features                                                   & CNN                       & \textit{EER = 12.3\% \cite{2019_Arxiv_GANRemoval_Tolosana}}             & \textit{100K-Faces (StyleGAN)}   \\ \hline \hline

\begin{tabular}[c]{@{}c@{}}Yu \textit{et al.} (2019)\\ \cite{yu2018attributing}\end{tabular}             & Deep Learning Features                                                    & CNN                       & Acc. = 99.5\%            & \begin{tabular}[c]{@{}c@{}}Own\\ (ProGAN, SNGAN,\\CramerGAN, MMDGAN)\end{tabular}   \\ \hline

\begin{tabular}[c]{@{}c@{}}Marra \textit{et al.} (2019)\\ \cite{WIFS2019_Luisa}\end{tabular}     & Deep Learning Features        & CNN + Incremental Learning & Acc. = 99.3\%                                                                        & \begin{tabular}[c]{@{}c@{}}Own\\ (CycleGAN, ProGAN, \\Glow, StarGAN, StyleGAN)\end{tabular} \\  \hline

\begin{tabular}[c]{@{}c@{}}Dang \textit{et al.} (2020)\\ \cite{Jain2019facialManipulation}\end{tabular}     & Deep Learning Features & CNN + Attention Mechanism & \textbf{\begin{tabular}[c]{@{}c@{}}AUC = 100\%\\ EER = 0.1\%\end{tabular}} & \textbf{DFFD (ProGAN, StyleGAN)}    \\ \hline

\multirow{2}{*}{\begin{tabular}[c]{@{}c@{}}Neves \textit{et al.} (2020)\\ \cite{2019_Arxiv_GANRemoval_Tolosana}\end{tabular}} & \multirow{2}{*}{Deep Learning Features}                                 & \multirow{2}{*}{CNN}      & \textbf{EER = 0.3\%}              & \textbf{100K-Faces (StyleGAN)}                 \\ \cline{4-5}
                                                                                            &                                                                         &                           & \textbf{EER = 4.5\%}             & \textbf{iFakeFaceDB} \\ \hline

\begin{tabular}[c]{@{}c@{}}Hulzebosch \textit{et al.} (2020)\\ \cite{realWorld_2020}\end{tabular}     & Deep Learning Features        & CNN, AE & Acc. = 99.8\%                                                                        & \begin{tabular}[c]{@{}c@{}}Own\\ (StarGAN, Glow, \\ProGAN, StyleGAN)\end{tabular} \\  \hline
                                                                                            
\end{tabular}
}
\end{table*}

Some authors propose to analyse the internal GAN pipeline in order to detect different artifacts between real and fake images. In~\cite{mccloskey2018detecting}, the authors hypothesised that the colour is markedly different between real camera images and fake synthesis images. They proposed a detection system based on colour features and a linear Support Vector Machine (SVM) for the final classification, achieving a final 70.0\% AUC for the best performance when evaluating with the NIST MFC2018 dataset~\cite{guan2019mfc}. 

Another interesting approach in this line was proposed in~\cite{wang2019fakespotter}. Wang \textit{et al.} conjectured that monitoring neuron behavior could also serve as an asset in detecting fake faces since layer-by-layer neuron activation patterns may capture more subtle features that are important for the facial manipulation detection system. Their proposed approach, named FakeSpoter, extracted as features neuron coverage behaviors of real and fake faces from deep face recognition systems (i.e., VGG-Face~\cite{vggface}, OpenFace~\cite{openface}, and FaceNet~\cite{facenet}), and then trained a SVM for the final classification. The authors tested their proposed approach using real faces from CelebA-HQ~\cite{pgan} and FFHQ~\cite{Karras_2019_CVPR} databases and synthetic faces created through InterFaceGAN~\cite{shen2019interpreting} and StyleGAN~\cite{Karras_2019_CVPR}, achieving for the best performance a final 84.7\% fake detection accuracy using the FaceNet model.

Better results have been recently reported in~\cite{convolutional_traces_2020}. The authors proposed a fake detection system based on the analysis of the convolutional traces. Features were extracted using the Expectation Maximization algorithm~\cite{moon1996expectation}. Popular classifiers such as \textit{k}-Nearest Neighbours (\textit{k}-NN), SVM, and Linear Discriminant Analysis (LDA) were used for the final detection. Their proposed approach was tested using fake images generated through AttGAN~\cite{he2019attgan}, GDWCT~\cite{cho2019image}, StarGAN~\cite{choi2018stargan}, StyleGAN, and StyleGAN2~\cite{karras2019analyzing}, achieving a final 99.81\% Acc. for the best performance.

Fake detection systems inspired in steganalysis have also been studied. Nataraj \textit{et al.} proposed in~\cite{nataraj2019detecting} a detection system based on a combination of pixel co-occurrence matrices and Convolutional Neural Networks (CNN). Their proposed approach was initially tested through a database of various objects and scenes created through CycleGAN~\cite{zhu2017unpaired}. Besides, the authors performed an interesting analysis to see the robustness of the proposed approach against fake images created through different GAN architectures (CycleGAN vs. StarGAN), with good generalisation results. This detection approach was implemented later on in~\cite{2019_Arxiv_GANRemoval_Tolosana} considering images from the 100K-Faces database, achieving an EER of 12.3\% for the best fake detection performance. This result is remarked in \textit{italics} in Table~\ref{table:relatedWorks_entireFaceSynthesis} to indicate that it was not provided in the original paper. 

Many studies have also focused on the detection of the special fingerprints inserted by GAN architectures using pure deep learning methods. Yu \textit{et al.} proposed in~\cite{yu2018attributing} an attribution network architecture to map an input image to its corresponding fingerprint image. Therefore, they learned a model fingerprint for each source (each GAN instance plus the real world), such that the correlation index between one image fingerprint and each model fingerprint serves as softmax logit for classification. Their proposed approach was tested using real faces from CelebA database~\cite{celeba} and synthetic faces created through different GAN approaches (ProGAN~\cite{pgan}, SNGAN~\cite{sngan}, CramerGAN~\cite{cramergan}, and MMDGAN~\cite{mmdgan}), achieving a final 99.5\% fake detection accuracy for the best performance. However, this approach seemed not to be very robust against unseen simple image perturbation attacks such as noise, blur, cropping or compression, unless the models were re-trained again.

Related to the unseen conditions just commented, Marra \textit{et al.} performed in~\cite{WIFS2019_Luisa} an interesting study in order to detect unseen types of fake generated data. Concretely, they proposed a multi-task incremental learning detection method in order to detect and classify new types of GAN generated images, without worsening the performance on the previous ones. Two different solutions regarding the position of the classifier were proposed based on the successful algorithm iCaRL for incremental learning~\cite{rebuffi2017icarl}: \textit{i)} Multi-Task MultiClassifier (MT-MC), and \textit{ii)} Multi-Task Single Classifier (MT-SC). Regarding the experimental framework, five different GAN approaches were considered in the study, CycleGAN~\cite{zhu2017unpaired}, ProGAN~\cite{pgan}, Glow~\cite{kingma2018glow}, StarGAN~\cite{choi2018stargan}, and StyleGAN~\cite{Karras_2019_CVPR}. Their proposed detection approach, based on the XceptionNet model, achieved promising results being able to correctly detect new GAN generated images.

Attention mechanisms have also been applied to further improve the training process of the detection systems. Dang \textit{et al.} carried out in~\cite{Jain2019facialManipulation} a complete analysis of different types of facial manipulations. They proposed to use attention mechanisms and popular CNN models such as XceptionNet and VGG16. For the entire face synthesis manipulation, the authors achieved a final 100\% AUC and around 0.1\% EER considering real faces from CelebA~\cite{celeba}, FFHQ~\cite{Karras_2019_CVPR}, and FaceForensics++~\cite{rossler2019faceforensics++} databases and fake images created through ProGAN~\cite{pgan} and StyleGAN~\cite{Karras_2019_CVPR} approaches. The impressive results achieved show the importance of novel attention mechanisms~\cite{vaswani2017attention}. 

Neves \textit{et al.} performed in~\cite{2019_Arxiv_GANRemoval_Tolosana} an in-depth experimental assessment of this type of facial manipulation considering different state-of-the-art detection systems and experimental conditions, i.e., controlled and in-the-wild scenarios. Four different fake databases were considered: \textit{i)} 150,000 fake faces collected online\footnote{\url{https://thispersondoesnotexist.com}} and based on StyleGAN architecture, \textit{ii)} the 100K-faces public database, \textit{iii)} 80,000 synthetic faces generated using ProGAN, and \textit{iv)} the iFakeFaceDB database, an improved version of previous fake databases in which the GAN-fingerprint information has been removed using the GANprintR approach. In controlled scenarios, they achieved similar results as the best previous studies (EER = 0.02\%). However, in more challenging scenarios in which images (real and fake) come from different sources (mismatch of datasets), a high degradation of the fake detection performance is observed. Finally, the results achieved over their public iFakeFaceDB database with an EER = 4.5\% for the best fake detectors remark how challenging is iFakeFaceDB even for the most advanced manipulation detection methods. Related to this enhanced fake content, Cozzolino \textit{et al.} proposed in~\cite{cozzolino2019spoc} a similar approach based on GAN to inject camera traces into synthetic images to spoof state-of-the-art fake detectors.

Similar to~\cite{2019_Arxiv_GANRemoval_Tolosana}, Hulzebosch \textit{et al.} have recently performed in~\cite{realWorld_2020} an in-depth analysis of this face manipulation considering different scenarios such as cross-model, cross-data, and post-processing. Fake detection approaches were based on the popular Xception network and ForensicTransfer~\cite{cozzolino2018forensictransfer}, which is an Autoencoder approach. In general, bad generalisation results were obtained under unseen scenarios, similar to~\cite{2019_Arxiv_GANRemoval_Tolosana}.
 
Finally, we also include for completeness some important references to other recent studies focused on the detection of general GAN-based image manipulations, not facial ones. In particular, we refer the reader to~\cite{huh2018fighting,zhou2018learning}.

\section{Identity Swap}\label{face_swapping}

\subsection{Manipulation Techniques and Public Databases}\label{subsec:faceSwapping_databases}
This is one of the most popular face manipulation research lines nowadays due to the great public concerns around DeepFakes~\cite{TED_news_concerns,BBC_doubleVideos}. It consists of replacing the face of one person in a video with the face of another person. Unlike the entire face synthesis manipulation, where manipulations are carried out at image level, in identity swap the goal is to generate realistic fake videos.

Since publicly available fake databases such as the UADFV database~\cite{li2018ictu}, up to the recent Celeb-DF and DFDC databases~\cite{li2019celebdf,dolhansky2019deepfake}, many visual improvements have been carried out, increasing the realism of fake videos. As a result, identity swap databases can be divided into two different generations. Table~\ref{table:databases_faceSwap} summarises the main details of each public database, grouped in each generation. As can be seen, in this type of facial manipulation both real and fake videos are usually included in the databases.

In this section, we first provide the main details of each database, to finally summarise at a higher level the key differences among the two generations.

Three different databases are grouped in the first generation. UADFV was one of the first public databases~\cite{li2018ictu}. This database comprises 49 real videos from Youtube, which were used to create 49 fake videos through the FakeApp mobile application\footnote{\url{https://www.malavida.com/en/soft/fakeapp/}}, swapping in all of them the original face with the face of Nicolas Cage. Therefore, only one identity is considered in all fake videos. Each video represents one individual, with a typical resolution of 294$\times$500 pixels, and 11.14 seconds on average.

Korshunov and Marcel introduced in~\cite{korshunov2018deepfakes} the DeepfakeTIMIT database. This database comprises 620 fake videos of 32 subjects from the VidTIMIT database~\cite{sanderson2009multi}. Fake videos were created using the public GAN-based face-swapping algorithm\footnote{\url{https://github.com/shaoanlu/faceswap-GAN}}. In that approach, the generative network is adopted from CycleGAN~\cite{zhu2017unpaired}, using the weights of FaceNet~\cite{facenet}. The method Multi-Task Cascaded Convolution Networks is used for more stable detections and reliable face alignment~\cite{zhang2016joint}. Besides, the Kalman filter is also considered to smooth the bounding box positions over frames and eliminate jitter on the swapped face. Regarding the scenarios considered in DeepfakeTIMIT, two different qualities are considered: \textit{i)} low quality (LQ) with images of 64$\times$64 pixels, and \textit{ii)} high quality (HQ) with images of 128$\times$128 pixels. Additionally, different blending techniques were applied to the fake videos regarding the quality level.

\begin{table}[t]
\centering
\caption{\textbf{Identity Swap:} Publicly available databases.}
\vspace{3mm}
\label{table:databases_faceSwap}
\scalebox{0.8}{
\begin{tabular}{ccc}
\multicolumn{3}{c}{\textbf{\nth{1} Generation}} \\ \hline
\textbf{Database}                                                              & \textbf{Real Videos} & \textbf{Fake Videos}                                                      \\ \hline
\begin{tabular}[c]{@{}c@{}}UADFV (2018)\\ \cite{li2018ictu}\end{tabular}              & 49 (Youtube)         & 49 (FakeApp)                                                              \\ \hline
\begin{tabular}[c]{@{}c@{}}DeepfakeTIMIT (2018)\\ \cite{korshunov2018deepfakes}\end{tabular}     & -                    & 620 (faceswap-GAN)                                                        \\ \hline
\begin{tabular}[c]{@{}c@{}}FaceForensics++ (2019)\\ \cite{rossler2019faceforensics++}\end{tabular}   & 1,000 (Youtube)       & \begin{tabular}[c]{@{}c@{}}1,000 (FaceSwap)\\ 1,000 (DeepFake)\end{tabular} \\
                                   &             &             \\ 
\multicolumn{3}{c}{\textbf{\nth{2} Generation}} \\ \hline
\textbf{Database}                                                              & \textbf{Real Videos} & \textbf{Fake Videos} \\ \hline
\begin{tabular}[c]{@{}c@{}}DeepFakeDetection (2019)\\ \cite{Google_DeepFakes}\end{tabular} & 363 (Actors)         & 3,068 (DeepFake)                                                           \\ \hline
\begin{tabular}[c]{@{}c@{}}Celeb-DF (2019)\\ \cite{li2019celebdf}\end{tabular}          & 890 (Youtube)        & 5,639 (DeepFake)                                                            \\ \hline
\begin{tabular}[c]{@{}c@{}}DFDC Preview (2019)\\ \cite{dolhansky2019deepfake}\end{tabular}     & 1,131 (Actors)        & 4,119 (Unknown)                                                            \\ \hline
\end{tabular}
}
\end{table}

One of the most popular databases in this type of facial manipulation is FaceForensics++~\cite{rossler2019faceforensics++}. This database was introduced early 2019 as an extension of the original FaceForensics database~\cite{rossler2018faceforensics}, which was focused only on expression swap. FaceForensics++ contains 1000 real videos extracted from Youtube. Regarding the identity swap fake videos, they were generated using both computer graphics and DeepFake approaches (i.e., learning approach). For the computer graphics approach, the authors considered the publicly available FaceSwap algorithm\footnote{\url{https://github.com/MarekKowalski/FaceSwap}} whereas for the DeepFake approach, fake videos were created through the DeepFake FaceSwap GitHub implementation\footnote{\url{https://github.com/deepfakes/faceswap}}. The FaceSwap approach consists of face alignment, Gauss Newton optimization and image blending to swap the face of the source person to the target person. The DeepFake approach, as indicated in~\cite{rossler2019faceforensics++}, is based on two autoencoders with a shared encoder that are trained to reconstruct training images of the source and the target face, respectively. A face detector is used to crop and to align the images. To create a fake image, the trained encoder and decoder of the source face are applied to the target face. The autoencoder output is then blended with the rest of the image using Poisson image editing~\cite{perez2003poisson}. Regarding the figures of the FaceForensics++ database, 1000 fake videos were generated for each approach. Later  on, a new dataset named DeepFakeDetection, grouped inside the \nth{2} generation due to its higher realism, was included in the FaceForensics++ framework with the support of Google ~\cite{Google_DeepFakes}. This dataset comprises 363 real videos from 28 paid actors in 16 different scenes. Additionally, 3068 fake videos are included in the dataset based on DeepFake FaceSwap GitHub implementation. It is important to remark that for both FaceForensics++ and DeepFakeDetection databases different levels of video quality are considered, in particular: \textit{i)} RAW (original quality), \textit{ii)} HQ (constant rate quantization parameter equal to 23), and \textit{iii)} LQ (constant rate quantization parameter equal to 40). This aspect simulates the video processing techniques usually applied in social networks.   

\begin{figure*}[!]
\begin{center}
   \includegraphics[width=\linewidth]{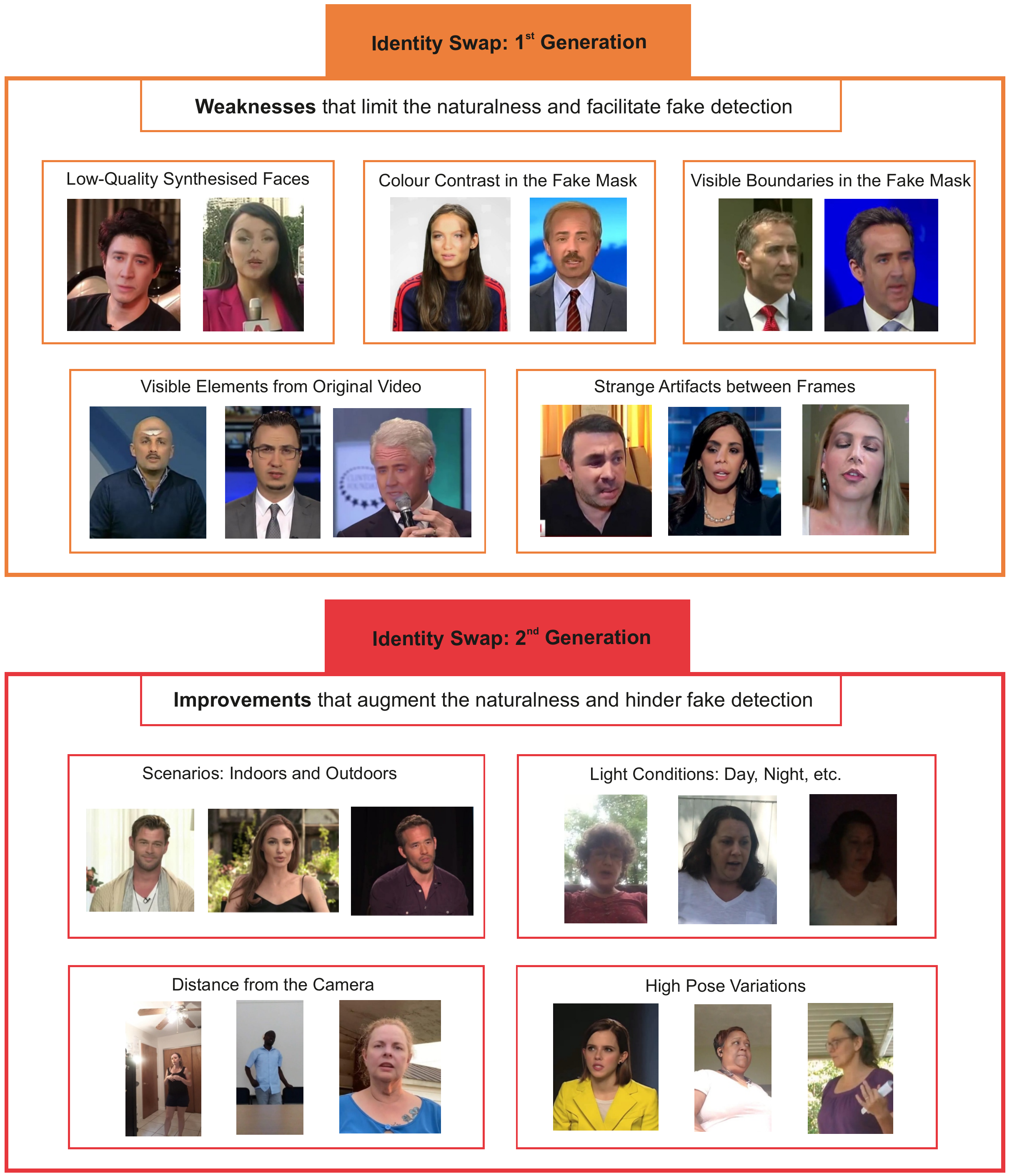}
\end{center}
   \caption{Graphical representation of the weaknesses present in identity swap databases of the \nth{1} generation and the improvements carried out in the \nth{2} generation, not only at visual level, but also in terms of variability (in-the-wild scenarios). Fake images are extracted from: UADFV and FaceForensics++ (\nth{1} generation)~\cite{li2018ictu,rossler2019faceforensics++}; Celeb-DF and DFDC (\nth{2} generation)~\cite{li2019celebdf,dolhansky2019deepfake}.}
\label{fig:deepFakes_generations}
\end{figure*}

Regarding the databases included in the \nth{2} generation, we highlight the recent Celeb-DF and DFDC databases released at the end of 2019. Li \textit{et al.} presented in~\cite{li2019celebdf} the Celeb-DF database. This database aims to provide fake videos of better visual qualities, similar to the popular videos that are shared on the Internet\footnote{\url{https://www.youtube.com/channel/UCKpH0CKltc73e4wh0\_pgL3g}}, in comparison to previous databases that exhibit low visual quality with many visible artifacts. Celeb-DF consists of 890 real videos extracted from Youtube, and 5,639 fake videos, which were created through a refined version of a public DeepFake generation algorithm, improving aspects such as the low resolution of the synthesised faces and colour inconsistencies. 

Facebook in collaboration with other companies and academic institutions such as Microsoft, Amazon, and the MIT launched at the end of 2019 a new challenge named the Deepfake Detection Challenge (DFDC)~\cite{dolhansky2019deepfake}. They first released a preview dataset consisting of 1,131 real videos from 66 paid actors, and 4,119 fake videos. Fake videos were generated using two different unknown approaches. The complete DFDC dataset was released later and comprises over 470 GB of content (real and fake)\footnote{\url{https://www.kaggle.com/c/deepfake-detection-challenge}}.

Finally, to conclude this section, we discuss at a higher level the key differences among fake databases from the \nth{1} and \nth{2} generations. In general, fake videos from the \nth{1} generation are characterised by: \textit{i)} low-quality synthesised faces, \textit{ii)} different colour contrast among the synthesised fake mask and the skin of the original face, \textit{iii)} visible boundaries of the fake mask, \textit{iv)} visible facial elements from the original video, \textit{v)} low pose variations, and \textit{vi)} strange artifacts among sequential frames. Also, they usually consider controlled scenarios in terms of camera position and light conditions. Many of these aspects have been successfully improved in databases of the \nth{2} generation, not only at visual level, but also in terms of variability (in-the-wild scenarios). For example, the recent DFDC database considers different acquisition scenarios (i.e., indoors and outdoors), light conditions (i.e., day, night, etc.), distances from the person to the camera, and pose variations, among others. Fig.~\ref{fig:deepFakes_generations} graphically summarises the weaknesses present in identity swap databases of the \nth{1} generation and the improvements carried out in the \nth{2} generation. Finally, it also interesting to remark the larger number of fake videos included in the databases of the \nth{2} generation.

\subsection{Manipulation Detection}\label{subsec:faceSwapping_stateoftheart}
The development of novel methods to detect identity swap manipulations is continuously evolving. Table~\ref{table:relatedWorks_faceSwapping} provides a comparison of the most relevant detection approaches in this area. For each study we include information related to the method, classifiers, best performance, and databases for research. We highlight \textbf{in bold} the best results achieved for each public database. It is important to remark that in some cases, different evaluation metrics are considered (e.g., AUC and EER), which complicates the comparison among studies. Finally, the results highlighted in \textit{italics} indicate the generalisation capacity of the detection systems against different unseen databases, i.e., those databases were not considered for training. These results have been extracted from~\cite{li2019celebdf} and were not included in the original publications.

\begin{table*}[!]
\centering
\caption{\textbf{Identity Swap:} Comparison of different state-of-the-art detection approaches. The best results achieved for each public database are remarked in \textbf{bold}. Results in \textit{italics} indicate that they were published in~\cite{li2019celebdf}, but not in the original work. FF++ = FaceForensics++, AUC = Area Under the Curve, Acc. = Accuracy, EER = Equal Error Rate, TCR = True Classification Rates.}
\vspace{3mm}
\label{table:relatedWorks_faceSwapping}
\scalebox{0.68}{
\begin{tabular}{ccccc}
\textbf{Study}                                                                                 & \textbf{Method}                       & \textbf{Classifiers}                & \textbf{Best Performance}                                                                       & \textbf{Databases}                                                                                                   \\ \hline
\begin{tabular}[c]{@{}c@{}}Korshunov and Marcel (2018)\\ \cite{korshunov2018deepfakes}\end{tabular}             & Audio-Visual Features & {\begin{tabular}[c]{@{}c@{}}PCA+RNN\\ PCA+LDA, SVM\end{tabular}}                  & \textbf{\begin{tabular}[c]{@{}c@{}}EER = 3.3\%\\ EER = 8.9\%\end{tabular}}                    & \textbf{\begin{tabular}[c]{@{}c@{}}DeepfakeTIMIT (LQ)\\ DeepfakeTIMIT (HQ)\end{tabular}}                             \\ \hline
\multirow{5}{*}{\begin{tabular}[c]{@{}c@{}}Matern \textit{et al.} (2019)\\ \cite{matern2019exploiting}\end{tabular}}   & \multirow{5}{*}{Visual Features}       & \multirow{5}{*}{\begin{tabular}[c]{@{}c@{}}Logistic Regression\\MLP\end{tabular}}        & AUC = 85.1\%                                                                                    & Own  \\ \cline{4-5}
                                                                                               &                                         &                                     & \textit{AUC = 70.2\%}                                                                           & \textit{UADFV}                                                                                                       \\ \cline{4-5}
                                                                                               &                                         &                                     & \textit{\begin{tabular}[c]{@{}c@{}}AUC = 77.0\%\\ AUC = 77.3\%\end{tabular}}                    & \textit{\begin{tabular}[c]{@{}c@{}}DeepfakeTIMIT (LQ)\\ DeepfakeTIMIT (HQ)\end{tabular}}                             \\ \cline{4-5}
                                                                                               &                                         &                                     & \textit{AUC = 78.0\%}                                                                           & \textit{FF++ / DFD}                                                                                                 
                                                                                               \\ \cline{4-5}
                                                                                               &                                         &                                     & \textit{AUC = 66.2\%}                                                                           & \textit{DFDC Preview}                                                                                                  \\ \cline{4-5}
                                                                                               &                                         &                                     & \textit{AUC = 55.1\%}                                                                           & \textit{Celeb-DF}                                                                                                    \\ \hline \hline

                                                                                               \multirow{4}{*}{\begin{tabular}[c]{@{}c@{}}Yang \textit{et al.} (2019)\\ \cite{yang2019exposing}\end{tabular}}     & \multirow{4}{*}{Head Pose Features}   & \multirow{4}{*}{SVM}                & AUC = 89.0\%                                                                                    & UADFV                                                                                                                \\ \cline{4-5}
                                                                                               &                                         &                                     & \textit{\begin{tabular}[c]{@{}c@{}}AUC = 55.1\%\\ AUC = 53.2\%\end{tabular}}                    & \textit{\begin{tabular}[c]{@{}c@{}}DeepfakeTIMIT (LQ)\\ DeepfakeTIMIT (HQ)\end{tabular}}                             \\ \cline{4-5}
                                                                                               &                                         &                                     & \textit{AUC = 47.3\%}                                                                           & \textit{FF++ / DFD}                                                                                                 
                                                                                               \\ \cline{4-5}
                                                                                               &                                         &                                     & \textit{AUC = 55.9\%}                                                                           & \textit{DFDC Preview}                                                                                                    \\ \cline{4-5}
                                                                                               &                                         &                                     & \textit{AUC = 54.6\%}                                                                           & \textit{Celeb-DF}                                                                                                    \\ \hline
                                                                                               \begin{tabular}[c]{@{}c@{}}Agarwal and Farid (2019)\\ \cite{2019_Agarwal}\end{tabular}     & Head Pose and Facial Features     & SVM          & AUC = 96.3\% & Own (FaceSwap, HQ)                                                                          \\ \hline \hline

\begin{tabular}[c]{@{}c@{}}Jung \textit{et al.} (2020)\\ \cite{eyeBlinking_2020}\end{tabular}     & Eye Blinking     & Distance          & Acc. = 87.5\% & Own                                                                           \\ \hline \hline

\multirow{4}{*}{\begin{tabular}[c]{@{}c@{}}Li \textit{et al.} (2019)\\ \cite{Li2019CVPR,li2019celebdf}\end{tabular}}       & \multirow{4}{*}{Face Warping Features} & \multirow{4}{*}{CNN}                & AUC = 97.7\% & UADFV                                                                                                       \\ \cline{4-5}
                                                                                               &                                         &                                     & \textbf{\begin{tabular}[c]{@{}c@{}}AUC = 99.9\%\\ AUC = 99.7\%\end{tabular}}                    & \textbf{\begin{tabular}[c]{@{}c@{}}DeepfakeTIMIT (LQ)\\ DeepfakeTIMIT (HQ)\end{tabular}}                             \\ \cline{4-5}
                                                                                               &                                         &                                     & \textit{AUC = 93.0\%}                                                                           & \textit{FF++ / DFD}                                                                                                 
                                                                                               \\ \cline{4-5}
                                                                                               &                                         &                                     & \textbf{\textit{AUC = 75.5\%}}                                                                           & \textit{DFDC Preview}                                                                                                  \\ \cline{4-5}
                                                                                               &                                         &                                     & \textit{AUC = 64.6\%}                                                                           & \textit{Celeb-DF}                                                                                                    \\ \hline \hline

\multirow{6}{*}{\begin{tabular}[c]{@{}c@{}}Afchar \textit{et al.} (2018)\\ \cite{afchar2018mesonet}\end{tabular}}   & \multirow{6}{*}{Mesoscopic Features}       & \multirow{6}{*}{CNN}                & Acc. = 98.4\%                                                                                   & Own  \\ \cline{4-5}
                                                                                               &                                         &                                     & \textit{AUC = 84.3\%}                                                                           & \textit{UADFV}                                                                                                       \\ \cline{4-5}
                                                                                               &                                         &                                     & \textit{\begin{tabular}[c]{@{}c@{}}AUC = 87.8\%\\ AUC = 68.4\%\end{tabular}}                    & \textit{\begin{tabular}[c]{@{}c@{}}DeepfakeTIMIT (LQ)\\ DeepfakeTIMIT (HQ)\end{tabular}}                             \\ \cline{4-5}
                                                                                               &                                         &                                     & \begin{tabular}[c]{@{}c@{}}Acc. $\simeq$ 90.0\%\\ Acc. $\simeq$ 94.0\%\\ Acc. $\simeq$ 98.0\%\end{tabular}           & \begin{tabular}[c]{@{}c@{}}FF++ (DeepFake, LQ)\\ FF++ (DeepFake, HQ)\\ FF++ (DeepFake, RAW)\end{tabular}          \\ \cline{4-5}
                                                                                               &                                         &                                     & \begin{tabular}[c]{@{}c@{}}Acc. $\simeq$ 83.0\%\\ Acc. $\simeq$ 93.0\%\\ Acc. $\simeq$ 96.0\%\end{tabular}           & \begin{tabular}[c]{@{}c@{}}FF++ (FaceSwap, LQ)\\ FF++ (FaceSwap, HQ)\\ FF++ (FaceSwap, RAW)\end{tabular}             
\\ \cline{4-5}
                                                                                               &                                         &                                     & \textit{AUC = 75.3\%}                                                                           & \textit{DFDC Preview}                                                                                                                                                                                                  \\ \cline{4-5}
                                                                                               &                                         &                                     & \textit{AUC = 54.8\%}                                                                           & \textit{Celeb-DF}                                                                                                    \\ \hline \hline

\multirow{4}{*}{\begin{tabular}[c]{@{}c@{}}Zhou \textit{et al.} (2018)\\ 
\cite{zhou2017two}\end{tabular}}      & \multirow{4}{*}{\begin{tabular}[c]{@{}c@{}}Steganalysis Features \\ + \\ Deep Learning Features\end{tabular}}           & \multirow{4}{*}{\begin{tabular}[c]{@{}c@{}}CNN\\SVM\end{tabular}} & \textit{AUC = 85.1\%}                                                                           & \textit{UADFV}                                                                                                       \\ \cline{4-5}
                                                                                               &                                         &                                     & \textit{\begin{tabular}[c]{@{}c@{}}AUC = 83.5\%\\ AUC = 73.5\%\end{tabular}}                    & \textit{\begin{tabular}[c]{@{}c@{}}DeepfakeTIMIT (LQ)\\ DeepfakeTIMIT (HQ)\end{tabular}}                             \\ \cline{4-5}
                                                                                               &                                         &                                     & \textit{AUC = 70.1\%}                                                                           & \textit{FF++ / DFD}
                                                                                               \\ \cline{4-5}
                                                                                               &                                         &                                     & \textit{AUC = 61.4\%}                                                                           & \textit{DFDC Preview}                                                                                                                                                                                                   \\ \cline{4-5}
                                                                                               &                                         &                                     & \textit{AUC = 53.8\%}                                                                  & \textit{Celeb-DF}                                                                                           \\ \hline
                                                                                               \multirow{2}{*}{\begin{tabular}[c]{@{}c@{}}R{\"o}ssler \textit{et al.} (2019)\\ \cite{rossler2019faceforensics++}\end{tabular}} & \multirow{2}{*}{\begin{tabular}[c]{@{}c@{}}Mesoscopic Features \\ Steganalysis Features \\ Deep Learning Features \end{tabular}} & \multirow{2}{*}{CNN}                & \begin{tabular}[c]{@{}c@{}}Acc. $\simeq$ 94.0\%\\ \textbf{Acc. $\simeq$ 98.0\%}\\ \textbf{Acc. $\simeq$ 100.0\%}\end{tabular} & \begin{tabular}[c]{@{}c@{}}FF++ (DeepFake, LQ)\\ \textbf{FF++ (DeepFake, HQ)}\\ \textbf{FF++ (DeepFake, RAW)}\end{tabular} \\ \cline{4-5}
                                                                                               &                                         &                                     & \begin{tabular}[c]{@{}c@{}}Acc. $\simeq$ 93.0\%\\ \textbf{Acc. $\simeq$ 97.0\%}\\ \textbf{Acc. $\simeq$ 99.0\%}\end{tabular}  & \begin{tabular}[c]{@{}c@{}}FF++ (FaceSwap, LQ)\\\textbf{FF++ (FaceSwap, HQ)}\\ \textbf{FF++ (FaceSwap, RAW)}\end{tabular}    \\ \hline \hline

\multirow{4}{*}{\begin{tabular}[c]{@{}c@{}}Nguyen \textit{et al.} (2019)\\ \cite{nguyen2019multi}\end{tabular}}  & \multirow{4}{*}{Deep Learning Features} & \multirow{4}{*}{AE + Multi-Task Learning}        & \textit{AUC = 65.8\%}                                                                           & \textit{UADFV}                                                                                                       \\ \cline{4-5}
                                                                                               &                                         &                                     & \textit{\begin{tabular}[c]{@{}c@{}}AUC = 62.2\%\\ AUC = 55.3\%\end{tabular}}                    & \textit{\begin{tabular}[c]{@{}c@{}}DeepfakeTIMIT (LQ)\\ DeepfakeTIMIT (HQ)\end{tabular}}                             \\ \cline{4-5}
                                                                                               &                                         &                                     & \textit{AUC = 76.3\%}                                                                           & \textit{FF++ / DFD}                                                                                                  \\ \cline{4-5}
                                                                                               &                                         &                                     & EER = 15.1\%                                                                                    & FF++ (FaceSwap, HQ)
                                                                                               \\ \cline{4-5}
                                                                                               &                                         &                                     & \textit{AUC = 53.6\%}                                                                                    & \textit{DFDC Preview}                                                                                                
                                                                                               \\ \cline{4-5}
                                                                                               &                                         &                                     & \textit{AUC = 54.3\%}                                                                                    & \textit{Celeb-DF}
                                                                                                \\ \hline
                                                                                               \multirow{4}{*}{\begin{tabular}[c]{@{}c@{}}Nguyen \textit{et al.} (2019)\\ \cite{nguyen2019use}\end{tabular}}  & \multirow{4}{*}{Deep Learning Features} & \multirow{4}{*}{Capsule Networks}        & \textit{AUC = 61.3\%}                                                                           & \textit{UADFV}                                                                                                       \\ \cline{4-5}
                                                                                               &                                         &                                     & \textit{\begin{tabular}[c]{@{}c@{}}AUC = 78.4\%\\ AUC = 74.4\%\end{tabular}}                    & \textit{\begin{tabular}[c]{@{}c@{}}DeepfakeTIMIT (LQ)\\ DeepfakeTIMIT (HQ)\end{tabular}}                             \\ \cline{4-5}
                                                                                               &                                         &                                     & \textit{AUC = 96.6\%}                                                                           & \textit{FF++ / DFD}                                                                                                  \\ \cline{4-5}
                                                                                               &                                         &                                     & \textit{AUC = 53.3\%}                                                                                    & \textit{DFDC Preview}
                                                                                               \\ \cline{4-5}
                                                                                               &                                         &                                     & \textit{AUC = 57.5\%}                                                                                    & \textit{Celeb-DF}                                                                                                                                                                                                 \\ \hline
\begin{tabular}[c]{@{}c@{}}Dang \textit{et al.} (2019)\\ \cite{Jain2019facialManipulation}\end{tabular}               & Deep Learning Features                  & CNN + Attention Mechanism                 & \begin{tabular}[c]{@{}c@{}}AUC = 99.4\%\\ EER = 3.1\%\end{tabular}                     & DFFD                                                                                                        \\ \hline
\begin{tabular}[c]{@{}c@{}}Dolhansky \textit{et al.} (2019)\\ \cite{dolhansky2019deepfake}\end{tabular}               & Deep Learning Features & CNN                                 & \textbf{\begin{tabular}[c]{@{}c@{}}\textbf{Precision = 93.0\%}\\ \textbf{Recall = 8.4\%}\end{tabular}}              & \textbf{DFDC Preview}                                                                                        \\ \hline 
\begin{tabular}[c]{@{}c@{}}Wang and Dantcheva (2020)\\ \cite{2019_Sabir}\end{tabular}     & {\begin{tabular}[c]{@{}c@{}}Deep Learning Features\end{tabular}}     & 3DCNN          & {\begin{tabular}[c]{@{}c@{}}\textbf{TCR = 95.13\%} \\\textbf{TCR = 92.25\%}\end{tabular}} & {\begin{tabular}[c]{@{}c@{}}\textbf{FF++ (DeepFake, LQ)}\\\textbf{FF++ (FaceSwap, LQ)}\end{tabular}} \\ \hline \hline

\begin{tabular}[c]{@{}c@{}}G\"uera and Delp (2018)\\ \cite{2018_David_AVSS}\end{tabular}     & {\begin{tabular}[c]{@{}c@{}}Image + Temporal Features \end{tabular}}    & CNN + RNN          & Acc. = 97.1\% & Own                                                                          \\ \hline
\begin{tabular}[c]{@{}c@{}}Sabir \textit{et al.} (2019)\\ \cite{2019_Sabir}\end{tabular}     & {\begin{tabular}[c]{@{}c@{}}Image + Temporal Features\end{tabular}}     & CNN + RNN          & {\begin{tabular}[c]{@{}c@{}}\textbf{AUC = 96.9\%} \\\textbf{AUC = 96.3\%}\end{tabular}} & {\begin{tabular}[c]{@{}c@{}}\textbf{FF++ (DeepFake, LQ)}\\\textbf{FF++ (FaceSwap, LQ)}\end{tabular}} \\ \hline \hline

\multirow{4}{*}{\begin{tabular}[c]{@{}c@{}}Tolosana \textit{et al.} (2020)\\ \cite{tolosana_FacialRegions}\end{tabular}}  & \multirow{4}{*}{Facial Regions Features} & \multirow{4}{*}{CNN}        & \textbf{AUC = 100.0\%}                                                                           & \textbf{UADFV}                                                                                                                                    \\ \cline{4-5}
                                                                                               &                                         &                                     & \textbf{AUC = 99.4\%}                                                                           & \textbf{FF++ (FaceSwap, HQ)}                                                                                                                                                                                                 \\ \cline{4-5}
                                                                                               &                                         &                                     & \textbf{AUC = 91.0\%}                                                                                    & \textbf{DFDC Preview}                                                                                                
                                                                                               \\ \cline{4-5}
                                                                                               &                                         &                                     & \textbf{AUC = 83.6\%}                                                                                    & \textbf{Celeb-DF}
                                                                                                \\ \hline
\end{tabular}
}
\end{table*}

The first studies in this area focused on the audio-visual artifacts existed in the \nth{1} generation of fake videos. Korshunov and Marcel evaluated in~\cite{korshunov2018deepfakes} baseline approaches based on the inconsistencies between lip movements and audio speech, as well as several variations of image-based systems often used in biometrics. For the first case, they considered Mel-Frequency Cepstral Coefficients (MFCCs) as audio features and distances between mouth landmarks as visual features. Principal Component Analysis (PCA) was then used to reduce the dimensionality of the blocks of features, and finally Recurrent Neural Networks (RNNs) based on Long Short-Term Memory (LSTM) to detect real of fake videos (based on~\cite{marcel2019EUSIPCO}). For the second case, they evaluated detection approaches based on: \textit{i)} raw faces as features, and \textit{ii)} image quality measures (IQM)~\cite{galbally13TIP}. In particular, they used a set of 129 features related to measures like signal to noise ratio, specularity, blurriness, etc. PCA with LDA, or SVM were considered for the final classification. Their proposed detection approach based on IQM+SVM provided the best results, with a final 3.3\% and 8.9\% EER for the LQ and HQ scenarios of the DeepfakeTIMIT database, respectively. 

In this line, Matern \textit{et al.} proposed in~\cite{matern2019exploiting} fake detection systems based on relatively simple visual aspects such as eye colour, missing reflections, and missing details in the eye and teeth areas. Two different classifiers were considered in this analysis: \textit{i)} a logistic regression model, and \textit{ii)} a Multilayer Perceptron (MLP)~\cite{goodfellow2016deep}. Their proposed approach was tested using a private database, achieving a final 85.1\% AUC for the MLP system. 

Fake detection systems based on facial expressions and head movements have also been proposed in the literature. Yang \textit{et al.} observes in~\cite{yang2019exposing} that some DeepFakes are created by splicing synthesised face regions into the original image, and in doing so, introducing errors that can be revealed when 3D head poses are estimated from the face images. Thus, they performed an study based on the differences between head poses estimated using a full set of facial landmarks (68 extracted from DLib~\cite{king2009dlib}) and those in the central face regions to differentiate DeepFakes from real videos. Once these features are extracted and normalised (mean and standard deviation), a SVM is considered for the final classification. Their proposed approach was originally evaluated with the UADFV database, achieving a final 89.0\% AUC. However, this pre-trained model (using UADFV database) seems not to generalise very well to other databases as depicted in Table~\ref{table:relatedWorks_faceSwapping}.

Another interesting approach in this line was proposed by Agarwal and Farid in~\cite{2019_Agarwal}. They proposed a detection system based on both facial expressions and head movements. For the feature extraction, the OpenFace2 toolkit was considered~\cite{baltrusaitis2018openface}, obtaining an intensity and occurrence for 18 different facial action units related to movements of facial muscles such as cheek raiser, nose wrinkle, mouth stretch, etc. Additionally, four features related to head movements were considered. As a result, each 10-second video clip is reduced to a feature vector of dimension 190 using the Pearson correlation to measure the linearity between features. Finally, the authors considered a SVM for the final classification. Regarding the experimental framework, the authors built their own database based on videos downloaded from YouTube of persons of interest talking in a formal setting, for example, weekly address, news interview, and public speech. In most videos the person is primarily facing towards the camera. Regarding the DeepFake videos, the authors trained one GAN per person based on faceswap-GAN\footnote{\url{https://github.com/shaoanlu/faceswap-GAN}}. Their proposed approach achieved a final 96.3\% AUC as the best fake detection performance, being robust against new contexts and manipulation techniques. 

Eye blinking~\cite{daza2020mebal} has also been studied to detect fake videos. In~\cite{eyeBlinking_2020}, the authors proposed an algorithm called DeepVision to analyse changes in the blinking patterns. Their approach was based on the fusion of Fast-HyperFace~\cite{ranjan2017hyperface} and Eye-Aspect-Ratio (EAR)~\cite{soukupova2016eye} to detect the face and obtain the eye aspect ratio. Finally, features based on blinking count and period were extracted to decide whether the video is real or fake. This approach achieved a final 87.5\% accuracy over a proprietary database. 

Another interesting research line is based on the detection of the artifacts included by the face manipulation pipeline. In~\cite{Li2019CVPR}, Li and Lyu hypothesised that some DeepFake algorithms can only create images of limited resolution, which need to be further warped to match the original faces in the source video. Such transforms leave distinctive artifacts in the resulting DeepFake videos. Thus, the authors proposed a detection system based on CNNs in order to detect the presence of such artifacts from the detected face regions and the surrounding areas. Four different CNN models were trained from scratch: VGG16~\cite{simonyan2014very}, ResNet50, ResNet101, and ResNet152~\cite{he2016deep}. Their proposed detection approach was tested using the UADFV and DeepfakeTIMIT databases, outperforming the state of the art for those databases. 

Li \textit{et al.} proposed later on in~\cite{li2019celebdf} an improved version of the work presented in~\cite{Li2019CVPR}. In this case, the authors included a new spatial pyramid pooling module to better handle the variations in the resolution~\cite{he2015spatial}. This detection approach was evaluated using different databases, achieving state-of-the-art results in some of them.

Approaches based on mesoscopic and steganalysis features have also been proposed in the literature. Afchar \textit{et al.} proposed in \cite{afchar2018mesonet} two different networks composed of few layers in order to focus on the mesoscopic properties of the images: \textit{i)} a CNN network comprised of 4 convolutional layers followed by a fully-connected layer (Meso-4), and \textit{ii)} a modification of Meso-4 consisted of a variant of the Inception module introduced in~\cite{szegedy2015going}, named MesoInception-4. Their proposed approach was originally tested against DeepFakes using a private database, achieving a 98.4\% of fake detection accuracy for the best performance. That pre-trained detection model was tested against unseen databases in~\cite{li2019celebdf}, proving to be a robust approach in some cases such as with FaceForensics++.

Zhou \textit{et al.} proposed a two-stream network for face manipulation detection. In particular, the authors considered a fusion of two streams: \textit{i)} a face classification stream based on the CNN GoogLeNet~\cite{szegedy2015going} to detect whether a face image is fake or not, and \textit{ii)} a path triplet stream that is trained using steganalysis features of images patches with a triplet loss, and a SVM for the classification. The initial system was trained to detect expression swap manipulations. Nevertheless, Li \textit{et al.} evaluated in~\cite{li2019celebdf} the generalisation capacity of the pre-trained model (trained using SwapMe app) to detect identity swap manipulations, resulting to be one the most robust approaches against the recent Celeb-DF database~\cite{li2019celebdf}.

An exhaustive analysis of different fake detection methods was carried out by R{\"o}ssler \textit{et al.} using FaceForensics++ database~\cite{rossler2019faceforensics++}. Five different detection systems were evaluated: \textit{i)} a CNN-based system trained through handcrafted steganalysis features~\cite{cozzolino2017recasting}, \textit{ii)} a CNN-based system whose convolution layers are specifically designed to suppress the high-level content of the image~\cite{bayar2016deep}, \textit{iii)} a CNN-based system with a global pooling layer that computes four statistics (mean, variance, maximum, and minimum)~\cite{rahmouni2017distinguishing}, \textit{iv)} the CNN MesoInception-4 detection system described in~\cite{afchar2018mesonet}, and finally \textit{v)} the CNN-based system XceptionNet~\cite{chollet2017xception} pre-trained using ImageNet database~\cite{deng2009imagenet} and re-trained for the face manipulation detection task. In general, the detection system based on XceptionNet architecture provided the best results in both types of manipulation methods, DeepFakes and FaceSwap. In addition, the detection systems were evaluated considering different video quality levels in order to simulate the video processing of many social networks. In this real scenario, the accuracy of all detection systems decreased when lowering the video quality, remarking how challenging is this task in real scenarios.

Recent deep learning methods considered in computer vision have been applied to further improve the detection of identity swap manipulations. In~\cite{nguyen2019multi}, Nguyen \textit{et al.} proposed a CNN system that uses multi-task learning to simultaneously detect fake videos and locate the manipulated regions. They considered a detection system based on an autoencoder. Concretely, they proposed to use a Y-shaped decoder in order to share valuable information between the classification, segmentation, and reconstruction tasks, improving the overall performance by reducing the loss. Their proposed approach was evaluated with the FaceSwap manipulation method for the FaceForensics++ database~\cite{rossler2018faceforensics}, achieving a best performance of 15.07\% EER, far from other detection approaches. In addition, this model seems not to generalise very well for other databases, with results below 80\% AUC. 

Later on, the same authors presented in~\cite{nguyen2019use} a new fake detection system based on the recent Capsule Networks. This approach uses fewer parameters than traditional CNN with similar performance~\cite{hinton2011transforming,sabour2017dynamic,Hinton_ICLRw}. The proposed detection system was originally evaluated using FaceForensics++ database with accuracies higher than 90\%. The same pre-trained detection model was tested against unseen databases in~\cite{li2019celebdf}, showing poor generalisation results, as it happens in most fake detection systems.

Attention mechanisms have also been applied to further improve the training process of the detection systems. Dang \textit{et al.} performed in~\cite{Jain2019facialManipulation} a thorough analysis of different face manipulations. They proposed a detection system based on CNN and attention mechanisms to process and improve the feature maps of the classifier model. Their proposed attention map can be implemented easily and inserted into existing backbone networks, through the inclusion of a single convolution layer, its associated loss functions, and masking the subsequent high-dimensional features. Their proposed detection approach was tested with the DFFD database (based on a combination of the previous FaceForensics++ databases and a collection of videos from the Internet). In particular, for identity swap detection, their proposed approach achieved an AUC of 99.43\% and EER of 3.1\%. Despite of the fact that it is difficult to provide a fair comparison among studies as different experimental protocols are considered, it is clear that their detection approach provides state-of-the-art results. 

In~\cite{dolhansky2019deepfake}, in addition to the description of the DFDC database, the authors provided baseline results using three simple detection systems: \textit{i)} a small CNN model composed of 6 convolution layers and 1 fully-connected layer to detect low-level image manipulations, \textit{ii)} an XceptionNet model trained using only face images, and \textit{iii)} an XceptionNet model trained using the full image. The detection system based on XceptionNet, considering only the face image (not the full image), provided the best results with 93.0\% precision and 8.4\% recall. 

Deep learning approaches based on 3DCNN were studied in~\cite{antitza_2020} in order to consider both spatial and motion information. In particular, the authors proposed fake detectors based on I3D~\cite{carreira2017quo} and 3D ResNet~\cite{hara2018can} approaches, achieving promising results on the low quality videos of FaceForensics++.

Detection systems based not only on features at image level, but also at temporal level, along the frames of the video, have also been studied in the literature. G\"uera and Delp proposed in~\cite{2018_David_AVSS} a temporal-aware pipeline to automatically detect fake videos. They considered a combination of CNNs and RNNs. For the CNN, the authors used InceptionV3~\cite{szegedy2016rethinking} pre-trained using ImageNet database~\cite{deng2009imagenet}. For the RNN system, they considered a LSTM model composed of one hidden layer with 2048 memory blocks. Finally, two fully-connected layers were included, providing the probabilities of the frame sequence being either real or fake. Their approach was evaluated using a proprietary database with a final 97.1\% accuracy. 

In this line, Sabir \textit{et al.} proposed a method to detect fake videos based on using the temporal information present in the stream~\cite{2019_Sabir}. The intuition behind this model is to exploit temporal discrepancies across frames. Thus, they considered a recurrent convolutional network similar to~\cite{2018_David_AVSS}, trained in this study end-to-end instead of using a pre-trained model. Their proposed detection approach was tested through FaceForensics++ database, achieving AUC results of 96.9\% and 96.3\% for the DeepFake and FaceSwap methods, respectively. Only the low-quality videos were considered in the analysis.

Finally, the discriminative power of each facial region for the detection of fake videos was studied in~\cite{tolosana_FacialRegions}. The authors considered a fake detection system based on XceptionNet. Databases from both \nth{1} and \nth{2} generations were considered in the experimental framework, concluding that poor fake detection results are achieved in the latest DeepFake video databases of the \nth{2} generation compared with the \nth{1} generation, with results of 91.0\% and 83.6\% AUC for the DFDC Preview and Celeb-DF databases, respectively. It is important to highlight that, contrary to~\cite{li2019celebdf}, a separate fake detection system was specifically trained for each database.

In conclusion, although many different approaches have been proposed in the literature, they all show poor generalisation results to unseen databases, as indicated in Table~\ref{table:relatedWorks_faceSwapping}. In addition, we also highlight the poor detection results achieved by most approaches on the DeepFake databases of the \nth{2} generation with results below 60\% AUC.

\section{Attribute Manipulation}\label{facial_attributes}

\subsection{Manipulation Techniques and Public Databases}\label{subsec:facialAttributes_databases}
This face manipulation consists of modifying in an image some attributes of the face such as the colour of the hair or the skin, the gender, the age, adding glasses, etc. Despite the success of GAN-based frameworks for general image translations and manipulations~\cite{choi2018stargan,suwajanakorn2017synthesizing,zhu2017unpaired,isola2017image,zhu2017toward,kim2017learning,bau2018gan}, and in particular for face attribute manipulations~\cite{choi2018stargan,Liu_2019_CVPR,perarnau2016invertible,li2016deep,shen2017learning,lample2017fader,xiao2018elegant,he2019attgan}, few databases are publicly available for research in this area, to the best of our knowledge. The main reason is that the code of most GAN approaches are publicly available, so researchers can easily generate their own fake databases as they like. Therefore, this section aims to highlight the latest GAN approaches in the field, from older to closer in time, providing also the link to their corresponding codes.

In~\cite{perarnau2016invertible}, the authors introduced the Invertible Conditional GAN (IcGAN)\footnote{\url{https://github.com/Guim3/IcGAN}} for complex image editing as the union of an encoder used jointly with a conditional GAN (cGAN)~\cite{mirza2014conditional}. This approach provides accurate results in terms of attribute manipulation. However, it seriously changes the face identity of the person. 

Lample \textit{et al.} proposed in~\cite{lample2017fader} an encoder-decoder architecture that is trained to reconstruct images by disentangling the salient information of the image and the attribute values directly in the latent space\footnote{\url{https://github.com/facebookresearch/FaderNetworks}}. However, as it happens with the IcGAN approach, the generated images may lack some details or present unexpected distortions.

An enhanced approach named StarGAN\footnote{\url{https://github.com/yunjey/stargan/blob/master/README.md}} was proposed in~\cite{choi2018stargan}. Before the StarGAN approach, many studies had shown promising results in image-to-image translations for two domains in general. However, few studies had focused on handling more than two domains. In that case a direct approach would be to build different models independently for every pair of image domains. StarGAN proposed a novel approach able to perform image-to-image translations for multiple domains using only a single model. The authors trained a conditional attribute transfer network via attribute classification loss and cycle consistency loss. Good visual results were achieved compared with previous approaches. However, it sometimes includes undesired modifications from the input face image such as the colour of the skin.

\begin{table*}[t]
\centering
\caption{\textbf{Attribute Manipulation:} Comparison of different state-of-the-art detection approaches. The best results achieved for each public database are remarked in \textbf{bold}. AUC = Area Under the Curve, Acc. = Accuracy, EER = Equal Error Rate.}
\vspace{3mm}
\label{table:relatedWorks_facialAttributes}
\scalebox{0.79}{
\begin{tabular}{ccccc}
\textbf{Study}                                                                    & \textbf{Method}    & \textbf{Classifiers}       & \textbf{Best Performance}                                                             & \textbf{Databases (Generation)}                                                                             \\ \hline
\begin{tabular}[c]{@{}c@{}}Wang \textit{et al.} (2019)\\ \cite{wang2019fakespotter}\end{tabular}      & GAN-Pipeline Features & SVM                        & Acc. = 84.7\%                                                                         & \begin{tabular}[c]{@{}c@{}}Own\\ (InterFaceGAN/StyleGAN)                                                                      \end{tabular} \\ \hline \hline

\begin{tabular}[c]{@{}c@{}}Nataraj \textit{et al.} (2019)\\ \cite{nataraj2019detecting}\end{tabular}   & Steganalysis Features         & CNN                        & Acc. = 99.4\%                                                                         & \begin{tabular}[c]{@{}c@{}}Own \\ (StarGAN/CycleGAN) \end{tabular}                                                                                                                                                                 \\ \hline \hline

\begin{tabular}[c]{@{}c@{}}Bharati \textit{et al.} (2016)\\ \cite{bharati2016detecting}\end{tabular}  & \begin{tabular}[c]{@{}c@{}} Deep Learning Features \\ (Face Patches) \end{tabular}         & RBM                        & \begin{tabular}[c]{@{}c@{}}Overall Acc. = 96.2\%\\ Overall Acc. = 87.1\%\end{tabular} & \begin{tabular}[c]{@{}c@{}}Own\\ (Celebrity Retouching,\\ ND-IIITD Retouching) \end{tabular} \\ \hline
\begin{tabular}[c]{@{}c@{}}Jain \textit{et al.} (2019)\\ \cite{jain2018detecting}\end{tabular}      & \begin{tabular}[c]{@{}c@{}} Deep Learning Features \\ (Face Patches) \end{tabular}         & CNN + SVM                  & \begin{tabular}[c]{@{}c@{}}Overall Acc. = 99.6\%\\ Overall Acc. = 99.7\%\end{tabular} & \begin{tabular}[c]{@{}c@{}}Own \\ (ND-IIITD Retouching, \\ StarGAN) \end{tabular}              \\ \hline \hline

\begin{tabular}[c]{@{}c@{}}Tariq \textit{et al.} (2018)\\ \cite{tariq2018detecting}\end{tabular}      & Deep Learning Features        & CNN                        &  \begin{tabular}[c]{@{}c@{}}AUC = 99.9\%\\ AUC = 74.9\%\end{tabular} & \begin{tabular}[c]{@{}c@{}}Own\\ (ProGAN, \\ Adobe Photoshop)  \end{tabular} \\ \hline
\begin{tabular}[c]{@{}c@{}}Dang \textit{et al.} (2019)\\ \cite{Jain2019facialManipulation}\end{tabular} & Deep Learning Features        & CNN +  Attention Mechanism & \textbf{\begin{tabular}[c]{@{}c@{}}AUC = 99.9\%\\ EER = 1.0\%\end{tabular}}                    & \textbf{DFFD (FaceApp/StarGAN)}                                                                         \\ \hline
\begin{tabular}[c]{@{}c@{}}Wang \textit{et al.} (2019)\\ \cite{wang2019detecting}\end{tabular}      & Deep Learning Features        & DRN                        & AP = 99.8\%                                                                           & \begin{tabular}[c]{@{}c@{}}Own \\ (Adobe Photoshop) \end{tabular}                                                                                         \\ \hline
\begin{tabular}[c]{@{}c@{}}Marra \textit{et al.} (2019)\\ \cite{WIFS2019_Luisa}\end{tabular}     & Deep Learning Features        & CNN + Incremental Learning & Acc. = 99.3\%                                                                        & \begin{tabular}[c]{@{}c@{}}Own \\ (Glow/StarGAN ) \end{tabular}                                                                                                                                                                                                                                             \\ \hline \hline

\begin{tabular}[c]{@{}c@{}}Zhang \textit{et al.} (2019)\\ \cite{zhang2019detecting}\end{tabular}     & Spectrum Domain Features     & GAN Discriminator          & Acc. = 100\%                                                                            & \begin{tabular}[c]{@{}c@{}}Own \\ (StarGAN/CycleGAN) \end{tabular}  \\ \hline \hline

\begin{tabular}[c]{@{}c@{}}Rathgeb \textit{et al.} (2020)\\ \cite{Rathgeb_2020_PRNU}\end{tabular}     & PRNU Features     & Score-Level Fusion          & EER = 13.7\%                                                                            & \begin{tabular}[c]{@{}c@{}}Own \\ (5 Public Apps) \end{tabular}  \\ \hline

\end{tabular}
}
\end{table*}

Almost at the same time He \textit{et al.} proposed in~\cite{he2019attgan} attGAN\footnote{\url{https://github.com/LynnHo/AttGAN-Tensorflow}}, a novel approach that removes the strict attribute-independent constraint from the latent representation, and just applies the attribute-classification constraint to the generated image to guarantee the correct change of the attributes. AttGAN provides state-of-the-art results on realistic attribute manipulation with other facial details well preserved. 

One of the latest approaches proposed in the literature is STGAN\footnote{\url{https://github.com/csmliu/STGAN}}~\cite{Liu_2019_CVPR}. In general, attribute manipulation can be tackled by incorporating an encoder-decoder or GAN. However, as commented Liu \textit{et al.}~\cite{Liu_2019_CVPR}, the bottleneck layer in the encoder-decoder usually provides blurry and low quality manipulation results. To improve this, the authors presented and incorporated selective transfer units with an encoder-decoder for simultaneously improving the attribute manipulation ability and the image quality. As a result, STGAN has recently outperformed the state of the art in attribute manipulation.

Despite of the fact that the code of most attribute manipulation approaches are publicly available, the lack of public databases and experimental protocols results crucial when comparing among different manipulation detection approaches. Otherwise, it is not possible to perform a fair comparison among studies. Up to now, to the best of our knowledge, the DFFD database~\cite{Jain2019facialManipulation} seems to be the only public database that considers this type of facial manipulations. This database comprises 18,416 and 79,960 fake images generated through FaceApp and StarGAN approaches, respectively.

\subsection{Manipulation Detection}\label{subsec:facialAttributes_stateoftheart}
Attribute manipulations have been originally studied in the field of face recognition in order to see how robust biometric systems are against physical factors such as plastic surgery, cosmetics, makeup or occlusions~\cite{kim2005effective,dantcheva2012can,kose2015facial,majumdar2019evading,christianIEEEAccess2019,9109348}. However, it has been the recent success of mobile applications such as FaceApp that has motivated the research community to detect digital face attribute manipulations. Table~\ref{table:relatedWorks_facialAttributes} provides a comparison of the most relevant approaches in this area. We include for each study information related to the method, classifiers, best performance, and databases for research.

Some authors propose to analyse the internal GAN pipeline to detect different artifacts between real and manipulated images. Similar to the entire face synthesis manipulations, Wang \textit{et al.} conjectured in~\cite{wang2019fakespotter} that monitoring neuron behavior could also serve as an asset in detecting fake faces since layer-by-layer neuron activation patterns may capture more subtle features that are important for the facial manipulation detection system. Their proposed approach, named FakeSpoter, extracted as features neuron coverage behaviors of real and fake faces from deep face recognition systems (VGG-Face~\cite{vggface}, OpenFace~\cite{openface}, and FaceNet~\cite{facenet}), and then trained a SVM for the final classification. The authors tested their proposed approach using real faces from CelebA-HQ~\cite{pgan} and FFHQ~\cite{Karras_2019_CVPR} databases and synthetic faces created through InterFaceGAN~\cite{shen2019interpreting} and StyleGAN~\cite{Karras_2019_CVPR}, achieving for the best performance a final 84.7\% manipulation detection accuracy using the FaceNet model.

Fake detection systems inspired in steganalysis have also been studied. As described in Sec. \ref{EntireFaceDet} for the entire face synthesis, Nataraj \textit{et al.} proposed in~\cite{nataraj2019detecting} a detection system based on the combination of pixel co-occurrence matrices and CNN. They created a new fake dataset based on attribute manipulations using the StarGAN approach~\cite{choi2018stargan} trained through the CelebA database~\cite{celeba}, achieving a final 99.4\% accuracy for the best result. 

Many studies have also focused on pure deep learning methods, either feeding the networks with face patches or with the complete face. In~\cite{bharati2016detecting}, Bharati \textit{et al.} proposed a deep learning approach based on a Restricted Boltzmann Machine (RBM) in order to detect digital retouching of face images. The input of the detection system consisted of face patches in order to learn discriminative features to classify each image as original or retouched. Regarding the databases, the authors generated two fake databases from the original ND-IIITD database (collection B~\cite{flynn2003assessment}) and a set of celebrity facial images downloaded from the Internet. Fake images were generated using the professional software PortraitPro Studio Max\footnote{\url{https://www.anthropics.com/portraitpro/}}, considering aspects such as skin texture, shape of eyes, nose, lips and overall face, prominence of smile, lip shape, and eye colour. Their proposed approach achieved overall accuracies for manipulation detection of 96.2\% and 87.1\% for the celebrity and ND-IIITD retouching databases, respectively. 

A similar approach based on non-overlapping face patches was presented in~\cite{jain2018detecting}. Jain \textit{et al.} proposed a CNN feature extractor composed of 6 convolutional layers and 2 fully-connected layers. Additionally, residual connections were considered inspired by a ResNet architecture~\cite{he2016deep}. Finally, a SVM was used for the final classification. Regarding the experimental framework, the ND-IIITD retouched database presented in~\cite{bharati2016detecting} was considered. Additionally, the authors considered fake images created through the StarGAN approach~\cite{choi2018stargan}, trained using the CelebA database~\cite{celeba}. In general, good detection results were achieved in both manipulation approaches, achieving almost 100\% manipulation detection accuracy.

Deep learning methods based on the complete face have been further studied in the literature, achieving in general very good results. Tariq \textit{et al.} evaluated in~\cite{tariq2018detecting} the use of different CNN architectures such as VGG16~\cite{vggface}, VGG19~\cite{vggface}, ResNet~\cite{he2016deep}, or XceptionNet~\cite{chollet2017xception}, among others. For the real face images, the CelebA database~\cite{celeba} was used. Regarding the fake images, two different approaches were considered: \textit{i)} machine approaches based on GAN, in particular ProGAN~\cite{pgan}, and \textit{ii)} manual approach based on Adobe Photoshop CS6, including manipulations such as makeup, glasses, sunglasses, hair, and hats. For the experimental evaluation, different sizes of the images were considered (from 32$\times$32 to 256$\times$256 pixels). A final 99.99\% AUC was obtained for the machine-created scenario whereas for the human-created scenario this value decreased to a final 74.9\% AUC for the best CNN model. Thus, a high degradation of the manipulation detection performance was observed between machine- and human-created fake images. 

Attention mechanisms have also been applied to further improve the training process of the detection systems. As described in previous sections, Dang \textit{et al.} developed in~\cite{Jain2019facialManipulation} a system able to detect different types of fakes. They used attention mechanisms to process and improve the feature maps of CNN models. Regarding the attribute manipulations, two different approaches were considered: \textit{i)} fake images created through the public FaceApp software, with up to 28 different available filters considering aspects such as hair, age, glasses, beard, and skin colour, among others; and \textit{ii)} fake images created through the StarGAN approach~\cite{choi2018stargan}, with up to 40 different filters. Their proposed approach was tested using their novel database DFFD, achieving very good results close to 1.0\% EER (and 99.9\% of AUC).

Wang \textit{et al.} carried out in~\cite{wang2019detecting} an interesting research using publicly available commercial software from Adobe Photoshop (Face-Aware Liquify tool~\cite{adobetool}) in order to synthesise new faces, and also a professional artist in order to manipulate 50 real photographs. The authors began running a human study through Amazon Mechanical Turk (AMT), showing real and fake images to the participants and asking them to classify each image into one of the classes. The results achieved remark how challenging the task is for humans, with a final 53.5\% of accuracy, close to chance (50\%). After the human study, the authors proposed two different automatic models: \textit{i)} a global classification model based on Dilated Residual Networks (DRN) to predict whether the face has been warped or not, and \textit{ii)} a local warp predictor based on the optical flow field in order to identify where manipulation occurs, and reverse them. The PWC-Net approach proposed in~\cite{sun2018pwc} was considered to compute the flow from original to manipulated and vice versa. Performances of 99.8\% and 97.4\% for automatic and manual face synthesis manipulation were achieved. 

The work~\cite{WIFS2019_Luisa} by Marra \textit{et al.} also described in Sec. \ref{EntireFaceDet} was able to correctly perform discrimination when new GANs were presented to the network and achieved a 99.3\% accuracy for their proposed manipulation detection approach, based on the XceptionNet model.

A detection system based on features extracted from the spectrum domain, rather than the raw image pixels, was presented by Zhang \textit{et al.} in~\cite{zhang2019detecting}. Given an image as input, they applied a 2D DFT to each of the RGB channels, getting one frequency image per channel. Regarding the classifier, they proposed AutoGAN, which is a GAN simulator that can synthesise GAN artifacts in any image without needing to access any pre-trained GAN model. The generalisation capacity of their proposed approach was tested using unseen GAN models. In particular, StarGAN~\cite{choi2018stargan} and GauGAN~\cite{isola2017image} were considered in the evaluation. For the StarGAN approach, good detection results were achieved using the frequency domain (100\%). However, for the GauGAN approach, a high degradation of the system performance, 50\% accuracy, was observed. The authors claimed that this was produced due to the generator of the GauGAN is drastically different from the CycleGAN (used in training).

Finally, Rathgeb \textit{et al.} proposed in~\cite{Rathgeb_2020_PRNU} a detection system based on Photo Response Non-Uniformity (PRNU). Specifically, scores obtained from the analysis of spatial and spectral features extracted from PRNU patterns across image cells were fused. Their proposed approach was evaluated over a private database created using 5 different mobile applications, achieving an average 13.7\% EER in manipulation detection.

To summarise this section, we can see that the core of most attribute manipulation detection systems are based on deep learning technology, providing in general very good results close to 100\% accuracy, as indicated in Table~\ref{table:relatedWorks_facialAttributes}. This is mainly produced due to the GAN-fingerprint information present in fake images. However, as indicated in the entire face synthesis manipulation, recent studies have been proposed in the literature to remove such GAN fingerprints from the fake images while keeping very realistic appearance~\cite{2019_Arxiv_GANRemoval_Tolosana,cozzolino2019spoc}, which represent a challenge even for the most advanced manipulation detectors.

\section{Expression Swap}\label{facial_expression}

\subsection{Manipulation Techniques and Public Databases}\label{subsec:facialExpression_databases}
This manipulation, also known as face reenactment,  consists of modifying the facial expression of the person. We focus on the most popular techniques Face2Face and NeuralTextures, which replace the facial expression of one person in a video with the facial expression of another person (also in a video). To the best of our knowledge, the only available database for research in this area is FaceForensics++~\cite{rossler2019faceforensics++}, an extension of FaceForensics~\cite{rossler2018faceforensics}. 

Initially, the FaceForensics database was focused on the Face2Face approach~\cite{thies2016face2face}. This is a computer graphics approach that transfers the expression of a source video to a target video while maintaining the identity of the target person. This was carried out through manual keyframe selection. Concretely, the first frames of each video were used to obtain a temporary face identity (i.e., a 3D model), and track the expression over the remaining frames. Then, fake videos were generated by transferring the source expression parameters of each frame (i.e., 76 Blendshape coefficients) to the target video. Later on, the same authors presented in FaceForensics++ a new learning approach based on NeuralTextures~\cite{thies2019deferred}. This is a rendering approach that uses the original video data to learn a neural texture of the target person, including a rendering network. In particular, the authors considered in their implementation a patch-based GAN-loss as used in Pix2Pix~\cite{isola2017image}. Only the facial expression corresponding to the mouth was modified. It is important to remark that all data is available on the FaceForensics++ GitHub\footnote{\url{https://github.com/ondyari/FaceForensics}}. In total, there are 1,000 real videos extracted from Youtube. Regarding the manipulated videos, 2,000 fake videos are available (1,000 videos for each considered fake approach). In addition, it is important to highlight that different video quality levels are considered, in particular: \textit{i)} RAW (original quality), \textit{ii)} HQ (constant rate quantization parameter equal to 23), and \textit{iii)} LQ (constant rate quantization parameter equal to 40). This aspect simulates the video processing techniques usually applied in social networks.

In addition to the Face2Face and NeuralTexture techniques considered in expression swap manipulations at video level, different approaches have been recently proposed to change the facial expression in both images and videos. A very popular approach was presented in~\cite{averbuch2017bringing}. Averbuch-Elor \textit{et al.} proposed a technique to automatically animate a still portrait using a video of a different subject, transferring the expressiveness of the subject of the video to the target portrait. Unlike Face2Face and NeuralTexture approaches that require videos from both input and target faces, in~\cite{averbuch2017bringing} just an image of the target is needed. In this line, a recent approach was recently presented in~\cite{zakharov2019few}, providing very good results in both one-shot and few-shot learning.

Finally, we also highlight other popular approaches at image level. For example, mobile applications such as FaceApp\footnote{\url{https://apps.apple.com/gb/app/faceapp-ai-face-editor/id1180884341}} allow to easily change the level of smiling, from happier to angrier. These approaches are based on current GAN architectures. For example, Choi \textit{et al.} showed in~\cite{choi2018stargan} the potential of StarGAN to change an input image to different expression levels such as angry, happy, neutral, sad, surprised, and fearful. Other recent GAN approaches that improve both the image quality of the fake images and the control editing of the parameters are InterFaceGAN~\cite{shen2019interpreting}, UGAN~\cite{zhu2019ugan}, STGAN~\cite{Liu_2019_CVPR}, and AttGAN~\cite{he2019attgan}.

\begin{table*}[t]
\centering
\caption{\textbf{Expression Swap:} Comparison of different state-of-the-art detection approaches. The best results achieved for each public database are remarked in \textbf{bold}. FF++ = FaceForensics++, AUC = Area Under the Curve, Acc. = Accuracy, EER = Equal Error Rate, TCR = True Classification Rate.}
\vspace{3mm}
\label{table:relatedWorks_facialExpression}
\scalebox{0.82}{
\begin{tabular}{ccccc}
\textbf{Study}                                                                            & \textbf{Method}                                                                     & \textbf{Classifiers}         & \textbf{Best Performance}                                         & \textbf{Databases (Generation)}         \\ \hline
\begin{tabular}[c]{@{}c@{}}Matern \textit{et al.} (2019)\\ \cite{matern2019exploiting}\end{tabular}                   & Visual Features & Logistic Regression, MLP                  & AUC = 86.6\%                                                      & FF++ (Face2Face, RAW)      \\ \hline \hline

\multirow{6}{*}{\begin{tabular}[c]{@{}c@{}}Afchar \textit{et al.} (2018)\\ \cite{afchar2018mesonet}\end{tabular}}  & \multirow{6}{*}{Mesoscopic Features}                                                     & \multirow{6}{*}{CNN}         & Acc. = 83.2\%                                                     & FF++ (Face2Face, LQ)       \\
                                                                                          &                                                                                       &                              & Acc. = 93.4\%                                                     & FF++ (Face2Face, HQ)       \\
                                                                                          &                                                                                       &                              & Acc. = 96.8\%                                                     & FF++ (Face2Face, RAW)      \\ \cline{4-5}
                                                                                          &                                                                                       &                              & Acc. $\simeq$ 75\%                                                       & FF++ (NeuralTextures, LQ)  \\
                                                                                          &                                                                                       &                              & Acc. $\simeq$ 85\%                                                       & FF++ (NeuralTextures, HQ)  \\
                                                                                          &                                                                                       &                              & Acc. $\simeq$ 95\%                                                         & FF++ (NeuralTextures, RAW) \\ \hline \hline

\multirow{6}{*}{\begin{tabular}[c]{@{}c@{}}R{\"o}ssler \textit{et al.} (2019)\\  \cite{rossler2019faceforensics++}\end{tabular}} & \multirow{6}{*}{\begin{tabular}[c]{@{}c@{}}Mesoscopic Features\\ Steganalysis Features \\Deep Learning Features\end{tabular}} & \multirow{6}{*}{CNN}         & Acc. $\simeq$ 91\%                                                       & FF++ (Face2Face, LQ)       \\
                                                                                          &                                                                                       &                              & \textbf{Acc. $\simeq$ 98\%}                                                       & \textbf{FF++ (Face2Face, HQ)}       \\
                                                                                          &                                                                                       &                              & \textbf{Acc. $\simeq$ 100\%}                                                      & \textbf{FF++ (Face2Face, RAW)}      \\ \cline{4-5}
                                                                                          &                                                                                       &                              & Acc. $\simeq$ 81\%                                                       & FF++ (NeuralTextures, LQ)  \\
                                                                                          &                                                                                       &                              & \textbf{Acc. $\simeq$ 93\%}                                                       & \textbf{FF++ (NeuralTextures, HQ)}  \\
                                                                                          &                                                                                       &                              & \textbf{Acc. $\simeq$ 99\%}                                                       & \textbf{FF++ (NeuralTextures, RAW)}  \\ \hline \hline

\multirow{2}{*}{\begin{tabular}[c]{@{}c@{}}Nguyen \textit{et al.} (2019)\\ \cite{nguyen2019multi}\end{tabular}}  & \multirow{2}{*}{Deep Learning Features}                                                        & \multirow{2}{*}{Autoencoder} & EER = 7.1\%                                                       & FF++ (Face2Face, HQ)       \\
                                                                                          &                                                                                       &                              & EER = 7.8\%                                                       & FF++ (NeuralTextures, HQ)  \\ \hline 
\begin{tabular}[c]{@{}c@{}}Dang \textit{et al.} (2020)\\ \cite{Jain2019facialManipulation}\end{tabular}                & Deep Learning Features                                                                         & CNN + Attention Mechanism    & \textbf{\begin{tabular}[c]{@{}c@{}}AUC = 99.4\%\\ EER = 3.4\%\end{tabular}} & \textbf{FF++ (Face2Face, -)}     \\ \hline 

\begin{tabular}[c]{@{}c@{}}Wang and Dantcheva (2020)\\ \cite{2019_Sabir}\end{tabular}     & {\begin{tabular}[c]{@{}c@{}}Deep Learning Features\end{tabular}}     & 3DCNN          & {\begin{tabular}[c]{@{}c@{}}TCR = 90.27\% \\TCR = 80.5\%\end{tabular}} & {\begin{tabular}[c]{@{}c@{}}\textbf{FF++ (Face2Face, LQ)}\\\textbf{FF++ (NeuralTextures, LQ)}\end{tabular}} \\ \hline \hline

\begin{tabular}[c]{@{}c@{}}Sabir \textit{et al.} (2019)\\ \cite{2019_Sabir}\end{tabular}              & \begin{tabular}[c]{@{}c@{}}Image + Temporal Features\end{tabular}                & CNN + RNN                    & \textbf{Acc. = 94.3}                                                       & \textbf{FF++ (Face2Face, LQ)}      \\ \hline
\begin{tabular}[c]{@{}c@{}}Amerini \textit{et al.} (2019)\\ \cite{2019_amerini_ICCV}\end{tabular}            & Image + Temporal Features                                                           & CNN + Optical Flow           & Acc. = 81.6\%                                                    & FF++ (Face2Face, -)        \\ \hline
\end{tabular}
}
\end{table*}

\subsection{Manipulation Detection}\label{subsec:facialExpression_stateoftheart}
This section aims to provide an overview of the expression swap detectors at video level using the FaceForensics++ database, as this is the only publicly available database for research in this area, to the best of our knowledge. Manipulations at image level (not video) can be detected using the same approaches described in Sec.~\ref{EntireFaceDet} and~\ref{subsec:facialAttributes_stateoftheart}. 

Table~\ref{table:relatedWorks_facialExpression} provides a comparison of the most relevant approaches in the area of expression swap detection. For each study we include information related to the method, classifiers, best performance, and databases. We highlight in \textbf{bold} the best results achieved for the only public database, FaceForensics++. It is important to remark that in some cases, different evaluation metrics are considered (e.g., AUC and EER), which makes it difficult to perform a fair comparison among the studies.

Some of the following methods were already discussed in Sect. \ref{subsec:faceSwapping_stateoftheart} for identity swap. Here we summarise the results achieved by them in detecting expression swap manipulations.

Preliminary studies have focused on the visual features existed in fake videos such as the eye colour, missing reflections, etc. In~\cite{matern2019exploiting} by Matern \textit{et al.}, the proposed approach was tested using FaceForensics++, only the Face2Face manipulation technique, achieving a final 86.6\% AUC for the best performance. 

Approaches based on mesoscopic and steganalysis features have also been studied in the literature. In~\cite{afchar2018mesonet}, the proposed approach was tested using the Face2Face fake videos from the FaceForensics++ database~\cite{rossler2019faceforensics++}, achieving in general good results, especially for RAW-quality videos. The same approach was later on tested in~\cite{rossler2019faceforensics++} against NeuralTextures fake videos, obtaining lower accuracy results compared with Face2Face. 

Recent deep learning methods have also been applied with good results. In~\cite{rossler2019faceforensics++}, the detection system based on XceptionNet provided the best results in both Face2Face and NeuralTextures manipulations, close to 100\% on RAW quality. In addition, the detection systems were evaluated considering different video quality levels in order to simulate the video processing of many social networks. In this real scenario, the accuracy of all detection systems was degraded with the video quality, as it happens in identity swap manipulations.

In~\cite{nguyen2019multi}, the proposed approach based on multi-task learning was evaluated with the FaceForensics++ database. For the Face2Face method, a 7.1\% EER was achieved on HQ videos whereas for the NeuralTexture method, the EER increased a bit more to a final 7.8\% EER in manipulation detection.

Attention mechanisms have been recently proposed in~\cite{Jain2019facialManipulation} to further improve the training process. The proposed detection approach was tested using the DFFD database, which for the expression swap manipulation is based only on data from FaceForensics++ database. The proposed approach achieved an AUC = 99.4\% and EER = 3.4\%.

Deep learning approaches based on 3DCNN were studied in~\cite{antitza_2020} in order to consider both spatial and motion information. Similar to the identity swap manipulation, the authors proposed fake detectors based on I3D~\cite{carreira2017quo} and 3D ResNet~\cite{hara2018can} approaches, achieving promising results on the low quality videos of the FaceForensics++ database.

Another interesting line is based on the analysis of both image and temporal information. In \cite{2019_Sabir}, the proposed approach based on recurrent convolutional networks was tested using the FaceForensics++ database, achieving AUC results of 94.3\% for the Face2Face technique. Only the low-quality videos were considered in the analysis. Finally, in~\cite{2019_amerini_ICCV}, Amerini \textit{et al.} proposed the adoption of optical flow fields to exploit possible inter-frame dissimilarities, using the PWC-Net approach~\cite{sun2018pwc}. The optical flow is a vector field computed among consecutive frames to extract apparent motion in the scene. The use of this approach is motivated as fake videos should have unnatural optical flow due to the unusual movement of lips, eyes, etc. Preliminary results were obtained using both VGG16 and ResNet50 networks, obtaining an Acc. = 81.6\% for the best performance in manipulation detection. 
 
Finally, as stated previously, most of the approaches reported here for expression swap detection have also been used for identity swap detection as reviewed in Sec.~\ref{subsec:faceSwapping_stateoftheart}. In general, it seems that similar features can be learnt by the fake detectors to distinguish between real and fake content, achieving good results in both types of manipulations. We highlight the potential of novel techniques such as attention mechanisms to better guide the networks during the training process, as shown in~\cite{Jain2019facialManipulation}, achieving AUC results of 99.4\% for detecting both identity swap and expression swap manipulations.

\section{Other Face Manipulation Directions}\label{others_research_lines}
The four classes of face manipulation techniques described before are the ones that are receiving most attention in the last few years, but they do not perfectly represent all possible face manipulations. This section discusses some other challenging and dangerous approaches in face manipulation: face morphing, face de-identification, and face synthesis based on audio or text (i.e., audio-to-video and text-to-video).

\subsection{Face Morphing}

Face morphing is a type of face manipulation that can be used to create artificial biometric face samples that resemble the biometric information of two or more individuals \cite{wolberg1998image, MorphingReview19}. This means that the new morphed face image would be successfully verified against facial samples of these two or more individuals creating a serious threat to face recognition systems~\cite{gomez2017your,Marcel_morphing}. In this sense, face morphing is a different type of facial manipulation compared with the four main types covered in this survey. Also, it is worth noting that face morphing is mainly focused on creating fake samples at image level, not video such as identity swap manipulations.  

There has been recently a large amount of research in the field of face morphing. A very complete review of this field has been published by Scherhag \textit{et al.} \cite{MorphingReview19} in 2019 including both morphing techniques and also morphing attack detectors. Despite the large amount of publications, the research in this field is still in its infancy, with many open issues and challenges. It is important to highlight the lack of publicly available databases and benchmarks what makes it difficult to perform a fair comparison among studies. In order to overcome this aspect, Raja \textit{et al.} has recently presented an interesting framework for morphing attack detection~\cite{raja2020morphing}, including a publicly available database, evaluation platform, and benchmark\footnote{\url{https://biolab.csr.unibo.it/fvcongoing}}. The database comprises morphed and real images constituting 1,800 photographs of 150 subjects. Morphing images were generated using 6 different algorithms, presenting a wide variety of possible approaches.

Regarding the face morphing detectors, different approaches have been proposed in the literature based on different features, e.g.: the reduction of face details due to blending operations~\cite{kraetzer2017modeling}, Fourier spectrum of sensor pattern noise~\cite{zhang2018face}, differences between the facial landmarks~\cite{scherhag2018detecting, damer2018detecting}, and pure deep learning features~\cite{ferrara2019face, DiffMorphing20}. In addition, approaches based on face de-morphing have been studied in order to restore the accomplice's facial image~\cite{ferrara2017face, peng2019fd}.

\subsection{Face De-Identification}

The main goal of face de-identification (de-ID) is to remove the identity information present on a face image or video in order to preserve the privacy of the person~\cite{gross2006model}. This can be achieved in several ways. The simplest way can be just to obfuscate the face by blurring or pixelation (e.g., in Google Maps Street View). More sophisticated methods try to provide face images with different identities but maintaining all other factors (pose, expression, illumination, etc.) unaltered. Therefore, the concept of face de-ID is very general. One possible option to achieve face de-identification could be through face identity swap.

Earlier works in this area were based on applying face de-ID to still images. In \cite{DeID2009} Gross \textit{et al.} presented a multi-factor framework for de-ID, which combined linear, bilinear, and quadratic models. They showed their method was able to protect privacy while preserving data utility on an expression-variant face database. Recently, the developments of image synthesis methods based on generative deep neural networks, in particular GAN, have inspired new face de-ID methods such as \cite{meden2017face,DeIdCVPR17,DeIDCVPR2018,meden2018k,guo2018integrating,yi_lun_pan_2019}, which use synthesised faces to replace the original ones. Also, in~\cite{mirjalili2019flowsan}, the authors proposed the use of Semi-Adversarial Networks (SAN) to confound arbitrary face-based gender classifiers.

More recently, in \cite{DeIDgafni2019live} Gafni \textit{et al.} presented in 2019 a method that provides face de-ID with convincing performance even in unconstrained videos. Their approach is based on an adversarial autoencoder coupled with a trained face classifier. This way they can achieve a rich latent space, embedding both identity and expression information. Also, in \cite{DeIdentLi19} a new face de-ID method based on a deep transfer model was presented. This method treats the non-identity related facial attributes as the style of the original faces, and uses a trained facial attribute transfer model to extract and map them to different faces achieving very promising results both in images and videos.

Some other related studies in this area work directly over face representations or deep face models by eliminating there undesired or protected information like identity, gender, or facial expressions~\cite{alvi2018turning,morales2019sensitivenets,gong2019debface}. Once that protected information has been disentangled, a face image or video can then be generated based on the new representations originated in which the protected information has been eliminated, reduced, or obfuscated.

\subsection{Audio-to-Video and Text-to-Video}

A related topic to facial expression swap is the synthesis of video from audio or text. These types of video face manipulations are also known as lip-sinc deep fakes \cite{LipSin_CVPRw20}. Popular examples can be seen on the Internet\footnote{\url{https://www.youtube.com/watch?v=VWMEDacz3L4}} \footnote{\url{https://www.bbc.com/news/technology-48607673}}. 

Regarding the synthesis of fake videos from audio (audio-to-video), Suwajanakorn \textit{et al.} presented in \cite{suwajanakorn2017synthesizing} an approach to synthesise high quality videos of a person (Obama in this case) speaking with accurate lip sync. For this, they used as input to their approach many hours of previous videos of the person together with a new audio recording. In their approach they employed a recurrent neural network (based on LSTMs) to learn the mapping from raw audio features to mouth shapes. Then, based on the mouth shape at each frame, they synthesised high quality mouth texture, and composited it with 3D pose matching to create the new video to match the input audio track, producing photorealistic results. 

In \cite{Audio2Video_ijcai2019}, Song \textit{et al.} proposed an approach based on a novel conditional recurrent generation network that incorporates both image and
audio features in the recurrent unit for temporal dependency, and also a pair of spatial-temporal discriminators for better image/video quality. As a result, their approach can model both lip and mouth together with expression and head pose variations as a whole, achieving much more realistic results. The source code is publicly available in GitHub\footnote{\url{https://github.com/susanqq/Talking\_Face\_Generation}}. Also, in \cite{song2020everybody} Song \textit{et al.} presented a dynamic method not assuming a person-specific rendering network like in \cite{suwajanakorn2017synthesizing}. In their approach they are able to generate very realistic fake videos by carrying out a 3D face model reconstruction from the input video plus a recurrent network to translate the source audio into expression parameters. Finally, they introduced a novel video rendering network and a dynamic programming method to construct a temporally coherent and photo-realistic video. Video results are shown on the Internet\footnote{\url{https://wywu.github.io/projects/EBT/EBT.html}}. 

Another interesting approach was presented in~\cite{zhou2019talking}. Zhou \textit{et al.} proposed a novel framework called Disentangled Audio-Visual System (DAVS), which generates high quality talking face videos using disentangled audio-visual representation. Both audio and video speech information can be employed as input guidance. The source code is available in GitHub\footnote{\url{https://github.com/Hangz-nju-cuhk/Talking-Face-Generation-DAVS}}.	

Regarding the synthesis of fake videos from text (text-to-video), Fried \textit{et al.} proposed in \cite{Text2Video19} a method that takes as input a video of a person speaking and the desired text to be spoken, and synthesises a new video in which the person’s mouth is synchronised with the new words. In particular, their method automatically annotates an input talking-head video with phonemes, visemes, 3D face pose and geometry, reflectance, expression and scene illumination per frame. Finally, a recurrent video generation network creates a photorealistic video that matches the edited transcript. Examples of the fake videos generated with this approach are publicly available\footnote{\url{https://www.ohadf.com/projects/text-based-editing/}}.

To the best of our knowledge, there are no publicly available databases and benchmarks related to audio- and text-to-video fake detection content. Research on this topic is usually carried out through the synthesis of in-house data using publicly available implementations like the ones described in this section.

Recent studies have analysed how easy is to detect audio- and text-to-video fake content. In \cite{LipSin_CVPRw20}, Agarwal \textit{et al.} proposed a fake detection method that exploits the inconsistencies that exist between the dynamics of the mouth shape (visemes) and the spoken phoneme. They focused on some particular visemes in which the mouth must be completely closed and observed that this did not happen in many manipulated videos. Their proposed approach achieved good results, specially as the length of the video increases.

\section{Concluding Remarks}\label{conclusions}
Motivated by the ongoing success of digital face manipulations, specially DeepFakes, this survey provides a comprehensive panorama of the field, including details of up-to-date: \textit{i)} types of facial manipulations, \textit{ii)} facial manipulation techniques, \textit{iii)} public databases for research, and \textit{iv)} benchmarks for the detection of each facial manipulation group, including key results achieved by the most representative manipulation detection approaches.

Generally speaking, most current face manipulations seem easy to be detected under controlled scenarios, i.e., when fake detectors are evaluated in the same conditions they are trained for. This fact has been demonstrated in most of the benchmarks included in this survey, achieving very low error rates in manipulation detection. However, this scenario may not be very realistic as fake images and videos are usually shared on social networks, suffering from high variations such as compression level, resizing, noise, etc. Also, facial manipulation techniques are continuously improving. These factors motivate further research on the generalisation ability of the fake detectors against unseen conditions. This aspect has been preliminary studied in different works~\cite{2019_Arxiv_GANRemoval_Tolosana,yu2018attributing,WIFS2019_Luisa,realWorld_2020}. Future research could be in the line of the latest publications~\cite{one_class_2020,ABC_2020} as they do not require fake videos for training, providing a better generalisation ability to unseen attacks.

Fusion techniques, at a feature or score level, could provide a better adaptation of the fake detectors to the different scenarios~\cite{2018_INFFUS_MCSreview1_Fierrez,2018_INFFUS_MCSreview2_Fierrez,singh2019comprehensive}. In fact, different fake detection approaches are already based on the combination of different sources of information, e.g., Zhou \textit{et al.} proposed in~\cite{zhou2017two} a detection system based on the combination of steganalysis and pure deep learning features, whereas Rathgeb \textit{et al.} proposed in~\cite{Rathgeb_2020_PRNU} the combination of spatial and spectral features. Another two interesting fusion approaches have been recently presented in~\cite{fusion_PAs_2020,fusion_PAs_2020_2}, combining RGB, Depth, and InfraRed information to detect physical face attacks. Also, face weighting approaches have been proposed in order to detect fake videos using multiple frames~\cite{face_weighting_2020}. Finally, fusion of other sources of information such as the text, keystroke, or the audio that accompanies the videos when uploading them to social networks could be very valuable to improve the detectors~\cite{agrawal2017multimodal,shu2017fake,shu2019fakenewstracker,keystroke_fakeNews_2020}.

In addition to the traditional fake detectors based only on the image/video information, novel schemes should be studied in order to provide more robust tools. One example of this is the work presented by Tursman \textit{et al.} in~\cite{multiple_cameras_2020}. The authors proposed to detect fake content via social verification at capture time: the arbiters of truthfulness are a group of video cameras that synchronously capture a speaker, collectively reach consensus, and then sign their videos in real time as ``true''. Approaches like this one could further protect media content from attacks.

We highlight next the key aspects to improve and future trends to follow for each facial manipulation group:
\begin{itemize}
\item \textbf{Face Synthesis:} current manipulations are usually based on GAN architectures such as StyleGAN, providing very realistic images. Nevertheless, most detectors can easily distinguish between real and fake images, achieving accuracies close to 100\%. This is produced due to fake images are characterised by specific GAN fingerprints. But, what if we are able to remove those GAN fingerprints or add some noise patterns while keeping very realistic synthetic images? Recent approaches have focused on this research line, which represents a challenge even for the best manipulation detection systems~\cite{2019_Arxiv_GANRemoval_Tolosana,cozzolino2019spoc,farid_attacks_2020}.

\item \textbf{Identity Swap:} although many different approaches have been proposed in the literature, it is certainly difficult to decide which is the best one. This is produced due to many different factors. First, most approaches are trained for a specific database and compression level, achieving in general very good results. However, they all show poor generalisation results to unseen conditions. In addition, the fact that different metrics (i.e., Acc., AUC, EER, etc.) and experimental protocols are usually considered does not help to achieve fair comparisons among studies. All these aspects should be further considered to advance in the field. 

Furthermore, we want to highlight the detection results achieved in the latest DeepFake databases of the \nth{2} generation such as DFDC and Celeb-DF~\cite{li2019celebdf,dolhansky2019deepfake}. While fake detectors already achieve AUC results close to 100\% in databases of the \nth{1} generation such as UADFV and FaceForensics++~\cite{rossler2019faceforensics++,li2018ictu}, they all suffer from a high performance degradation on the latest ones, in particular for the Celeb-DF database with AUC results below 60\% in most cases. Therefore, more efforts are needed to further improve current fake detection systems, for example, through large-scale challenges and benchmarks such as the recent DFDC\footnote{\url{https://deepfakedetectionchallenge.ai/}}.

\item \textbf{Attribute Manipulation:} the same aspect highlighted for the face synthesis (GAN fingerprint removal) also applies here as most manipulations are based on GAN architectures. In addition, it is also interesting to remark the scarcity of public databases for research (only the DFFD database is publicly available~\cite{Jain2019facialManipulation}), and the lack of standard experimental protocols to perform fair comparisons among studies.

\item \textbf{Expression Swap:} contrary to the identity swap, which has rapidly evolved with the release of improved DeepFake databases, the only public database in expression swap is FaceForensics++, to the best of our knowledge. This database is characterised by visual artifacts that are easy to detect, achieving therefore AUC results close to 100\% in several fake detection approaches. We encourage researchers to generate and make public more realistic databases based on recent techniques~\cite{suwajanakorn2017synthesizing,averbuch2017bringing,song2020everybody}.
\end{itemize}

All these aspects, together with the development of improved GAN approaches and the recent DeepFake Detection Challenge (DFDC) will foster the new generation of realistic fake images/videos~\cite{karras2019analyzing} together with more advanced techniques for face manipulation detection.

\section*{Acknowledgments}
This work has been supported by projects: PRIMA (H2020-MSCA-ITN-2019-860315), TRESPASS-ETN (H2020-MSCA-ITN-2019-860813), BIBECA (MINECO/FEDER RTI2018-101248-B-I00), Bio-Guard (Ayudas Fundaci\'on BBVA a Equipos de Investigaci\'on Cient\'ifica 2017), and Accenture. Ruben Tolosana is supported by Consejer\'ia de Educaci\'on, Juventud y Deporte de la Comunidad de Madrid y Fondo Social Europeo.



%

%

{
\bibliographystyle{IEEEtran}
\bibliography{mybibfile}
}

%
%
\begin{IEEEbiography}[{\includegraphics[width=1in,height=1.25in,clip,keepaspectratio]{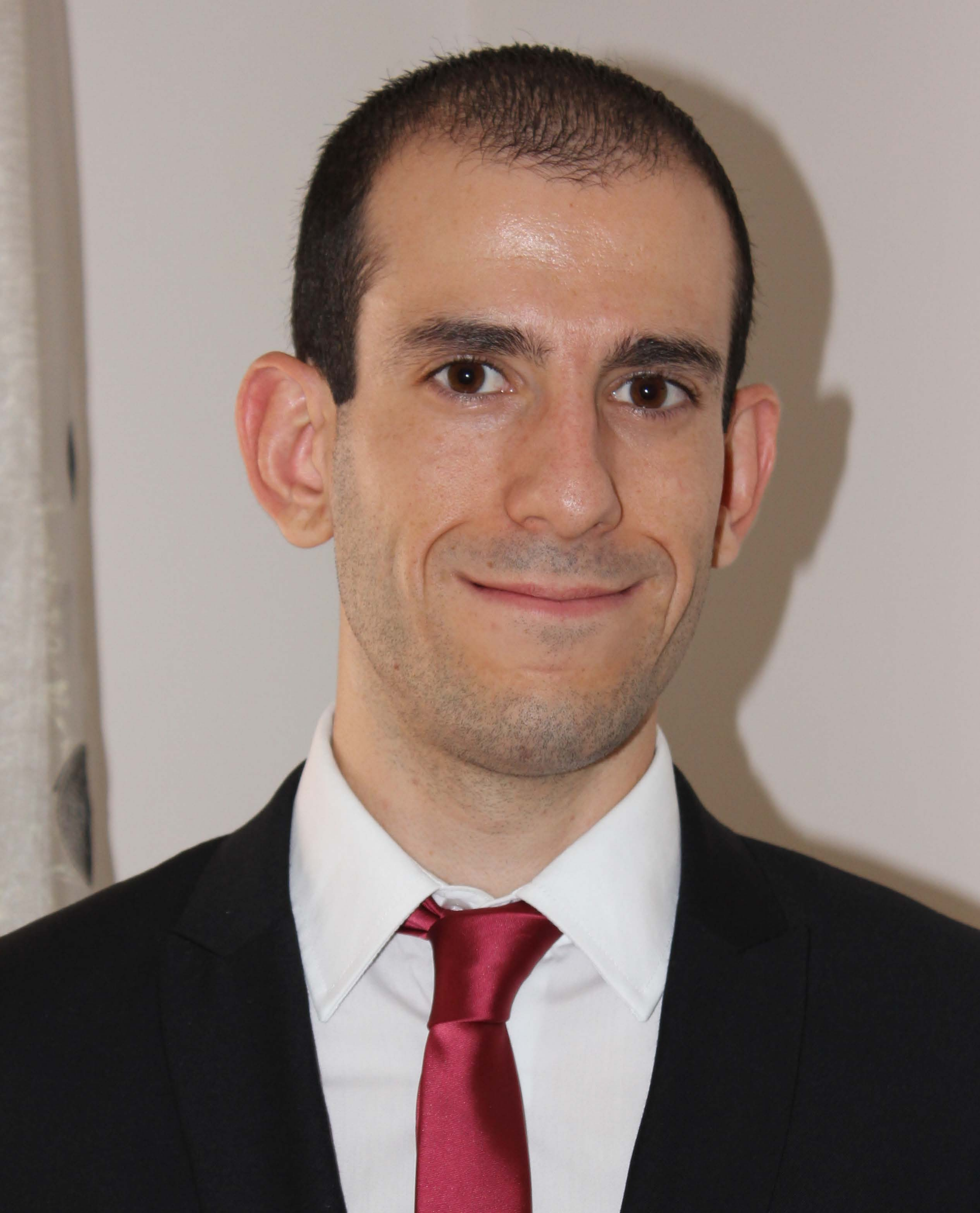}}]%
{Ruben Tolosana}
received the M.Sc. degree in Telecommunication Engineering, and his Ph.D. degree in Computer and Telecommunication Engineering, from Universidad Autonoma de Madrid, in 2014 and 2019, respectively. In April 2014, he joined the Biometrics and Data Pattern Analytics - BiDA Lab at the Universidad Autonoma de Madrid, where he is currently collaborating as a PostDoctoral researcher. Since then, Ruben has been granted with several awards such as the FPU research fellowship from Spanish MECD (2015), and the European Biometrics Industry Award (2018). His research interests are mainly focused on signal and image processing, pattern recognition, and machine learning, particularly in the areas of face manipulation, human-computer interaction and biometrics. He is author of several publications and also collaborates as a reviewer in many different high-impact conferences (e.g., ICDAR, IJCB, ICB, BTAS, EUSIPCO, etc.) and journals (e.g., IEEE TPAMI, TCYB, TIFS, TIP, ACM CSUR, etc.). Finally, he has participated in several National and European projects focused on the deployment of biometric security through the world.
\end{IEEEbiography}

\begin{IEEEbiography}[{\includegraphics[width=1in,height=1.25in,clip,keepaspectratio]{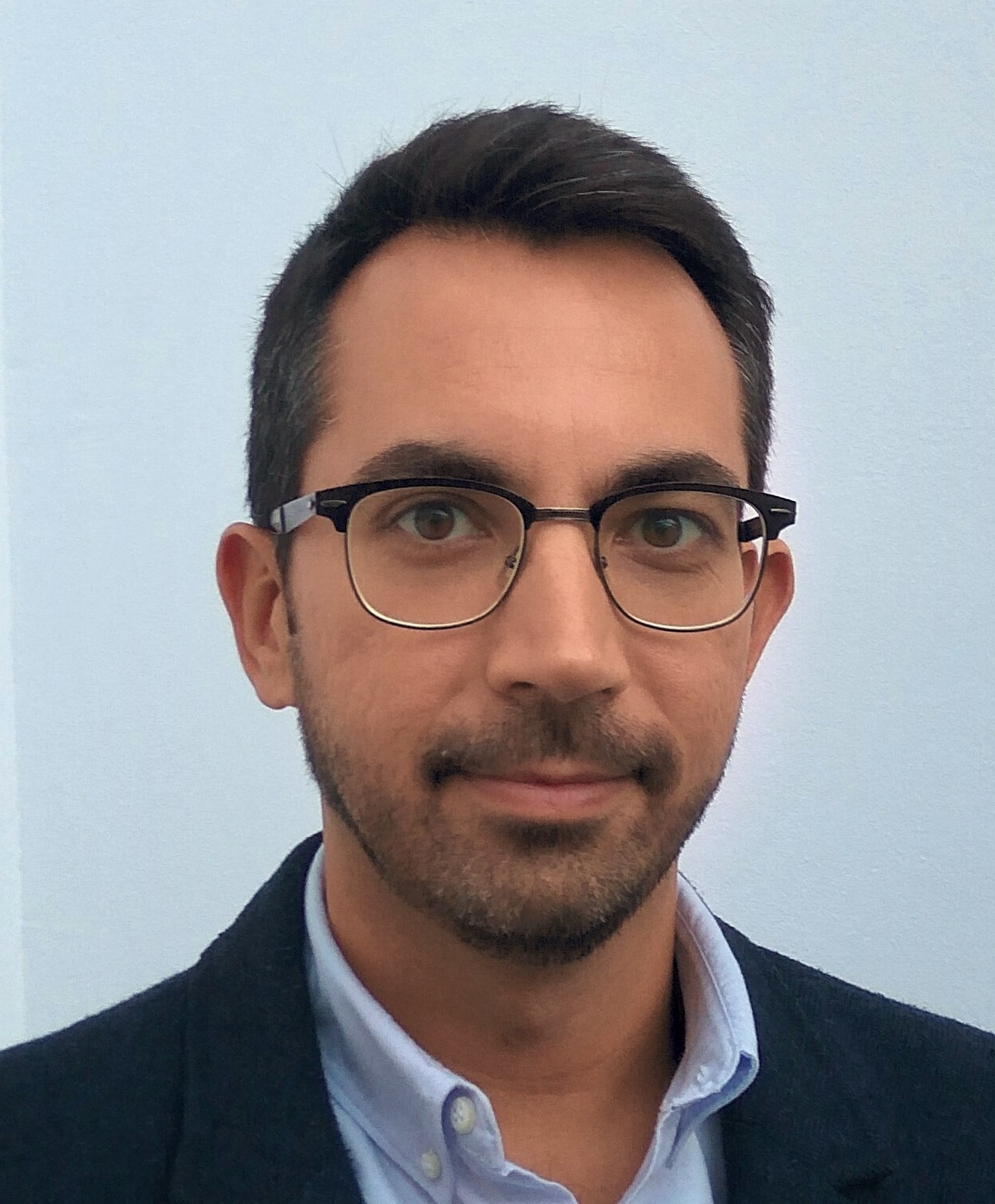}}]{Ruben Vera-Rodriguez} received the M.Sc. degree in telecommunications engineering from Universidad de Sevilla, Spain, in 2006, and the Ph.D. degree in electrical and electronic engineering from Swansea University, U.K., in 2010. Since 2010, he has been affiliated with the Biometric Recognition Group, Universidad Autonoma de Madrid, Spain, where he is currently an Associate Professor since 2018. His research interests include signal and image processing, pattern recognition, and biometrics, with emphasis on signature, face, gait verification and forensic applications of biometrics. He is actively involved in several National and European projects focused on biometrics. Ruben has been Program Chair for the IEEE 51st International Carnahan Conference on Security and Technology (ICCST) in 2017; and the 23rd Iberoamerican Congress on Pattern Recognition (CIARP 2018) in 2018.
\end{IEEEbiography}

\begin{IEEEbiography}[{\includegraphics[width=1in,height=1.25in,clip,keepaspectratio]{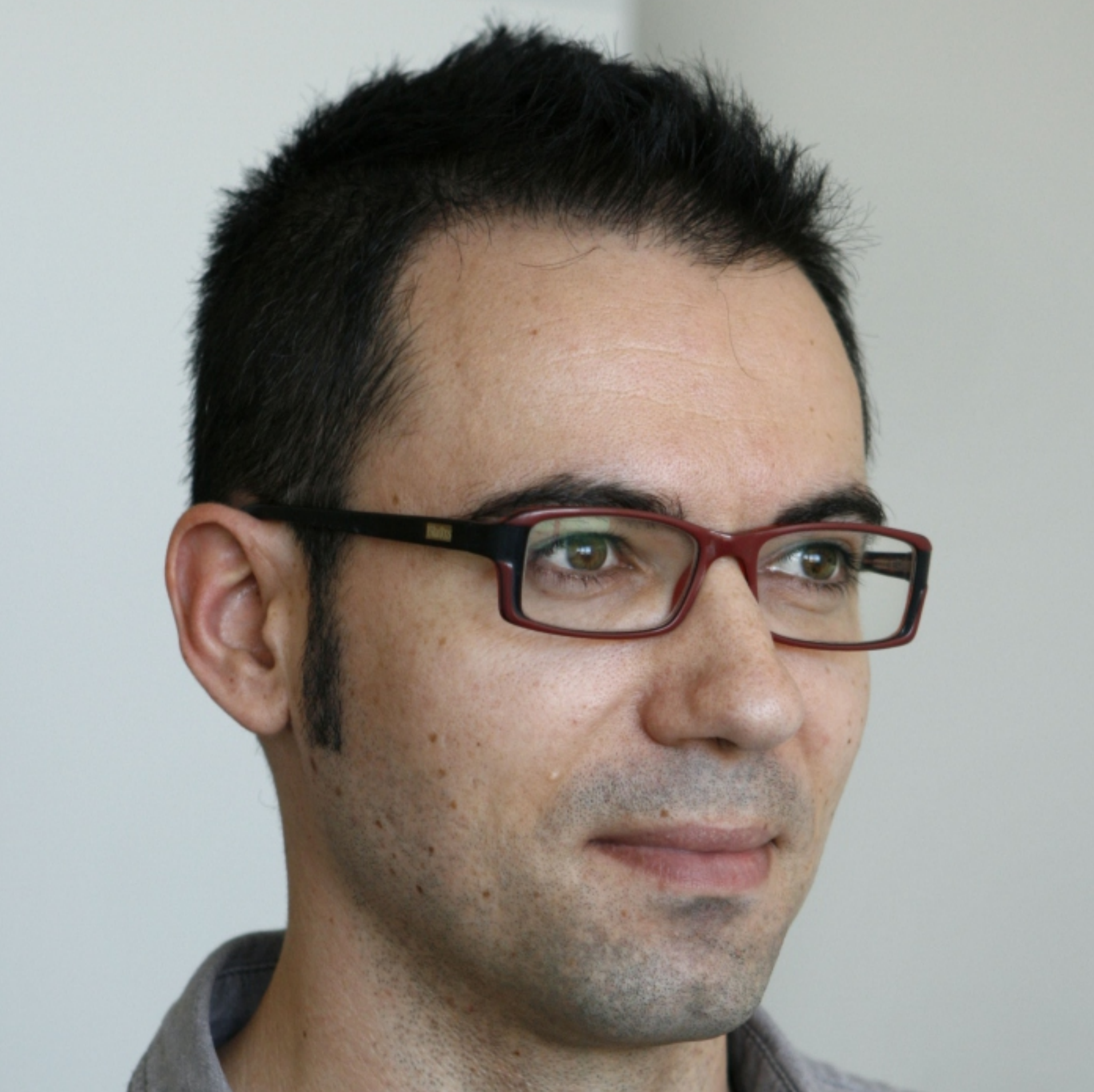}}]{Julian Fierrez} received the M.Sc. and Ph.D. degrees in telecommunications engineering from the Universidad Politecnica de Madrid, Spain, in 2001 and 2006, respectively. Since 2002, he has been with the Biometric Recognition Group, Universidad Politecnica de Madrid. Since 2004, he has been with the Universidad Autonoma de Madrid, where he is currently an Associate Professor. From 2007 to 2009, he was a Visiting Researcher with Michigan State University, USA, under a Marie Curie Fellowship. His research interests include signal and image processing, pattern recognition, and biometrics, with an emphasis on multibiometrics, biometric evaluation, system security, forensics, and mobile applications of biometrics. He has been actively involved in multiple EU projects focused on biometrics (e.g., TABULA RASA and BEAT), and has attracted notable impact for his research. He was a recipient of a number of distinctions, including the EAB European Biometric Industry Award 2006, the EURASIP Best Ph.D. Award 2012, the Miguel Catalan Award to the Best Researcher under 40 in the Community of Madrid in the general area of science and technology, and the 2017 IAPR Young Biometrics Investigator Award. He is an Associate Editor of the IEEE TRANSACTIONS ON INFORMATION FORENSICS AND SECURITY and the IEEE TRANSACTIONS ON IMAGE PROCESSING.
\end{IEEEbiography}

\begin{IEEEbiography}[{\includegraphics[width=1in,height=1.25in,clip,keepaspectratio]{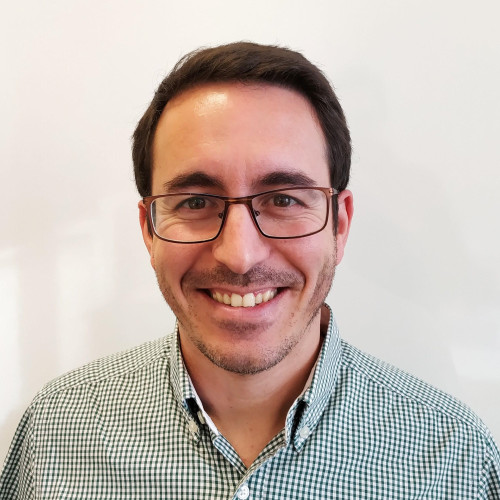}}]{Aythami Morales} received the M.Sc. degree in telecommunication engineering from the Universidad de Las Palmas de Gran Canaria in 2006 and the Ph.D. degree from La Universidad de Las Palmas de Gran Canaria in 2011. Since 2017, he is Associate Professor with the Universidad Autonoma de Madrid. He has conducted research stays at the Biometric Research Laboratory, Michigan State University, the Biometric Research Center, Hong Kong Polytechnic University, the Biometric System Laboratory, University of Bologna, and the Schepens Eye Research Institute. He has authored over 70 scientific articles published in international journals and conferences. He has participated in national and EU projects in collaboration with other universities and private entities, such as UAM, UPM, EUPMt, Indra, Union Fenosa, Soluziona, or Accenture. His research interests are focused on pattern recognition, computer vision, machine learning, and biometrics signal processing. He has received awards from the ULPGC, La Caja de Canarias, SPEGC, and COIT.
\end{IEEEbiography}

\begin{IEEEbiography}[{\includegraphics[width=1in,height=1.25in,clip,keepaspectratio]{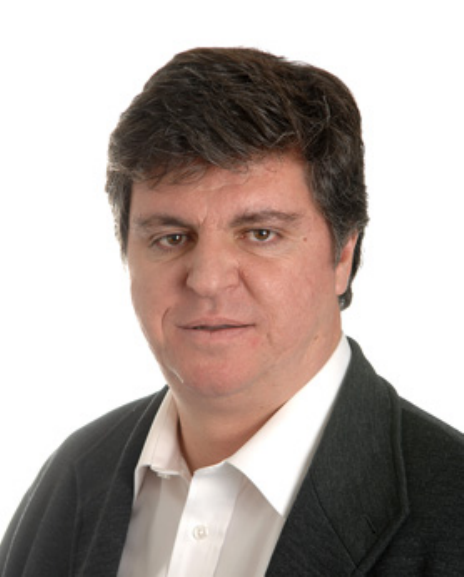}}]%
{Javier Ortega-Garcia}
received the M.Sc. degree in electrical engineering and the Ph.D. degree (cum laude) in electrical engineering from Universidad Politecnica de Madrid, Spain, in 1989 and 1996, respectively. He is currently a Full Professor at the Signal Processing Chair in Universidad Autonoma de Madrid - Spain, where he holds courses on biometric recognition and digital signal processing. He is a founder and Director of the BiDA-Lab, Biometrics and Data Pattern Analytics Group. He has authored over 300 international contributions, including book chapters, refereed journal, and conference papers. His research interests are focused on biometric pattern recognition (on-line signature verification, speaker recognition, human-device interaction) for security, e-health and user profiling applications. He chaired Odyssey-04, The Speaker Recognition Workshop, ICB-2013, the 6th IAPR International Conference on Biometrics, and ICCST2017, the 51st IEEE International Carnahan Conference on Security Technology.
\end{IEEEbiography}

\end{document}